\documentclass[letterpaper]{article}
\usepackage[preprint]{aaai2027}

\usepackage[hyphens]{url}
\usepackage{graphicx}
\usepackage{natbib}
\usepackage{caption}
\usepackage{booktabs}

\usepackage{amsmath,amssymb}
\usepackage{array}
\usepackage{multirow}
\usepackage{xcolor}
\usepackage{tikz}
\usetikzlibrary{arrows.meta,fit,positioning,shapes.geometric}

\usepackage[capitalise,noabbrev]{cleveref}

\definecolor{Ink}{HTML}{17202A}
\definecolor{Muted}{HTML}{5D6D7E}
\definecolor{Blue}{HTML}{2E6F9E}
\definecolor{BlueLight}{HTML}{EAF3F8}
\definecolor{Green}{HTML}{2F7D66}
\definecolor{GreenLight}{HTML}{E9F5F0}
\definecolor{Amber}{HTML}{A8660A}
\definecolor{AmberLight}{HTML}{FBF1DF}
\definecolor{Rule}{HTML}{CCD4DA}

\usepackage{enumitem}
\setlist{nosep,leftmargin=*}

\newcolumntype{L}[1]{>{\raggedright\arraybackslash}p{#1}}
\newcolumntype{C}[1]{>{\centering\arraybackslash}p{#1}}

\newcommand{\system}[1]{\textsc{#1}}
\newcommand{\agentic}{\system{Agentic RAG}}
\newcommand{\deepresearch}{\system{Deep Research Agent}}
\newcommand{\contextgraph}{\system{Context Graph Agent}}

\newcommand{\metric}[1]{\ensuremath{\mathrm{#1}}}

\tikzset{
  process/.style={draw=Rule, rounded corners=2pt, fill=white, align=center,
    inner xsep=5pt, inner ysep=4pt, font=\scriptsize},
  managedbox/.style={process, draw=Blue, fill=BlueLight},
  structurebox/.style={process, draw=Green, fill=GreenLight},
  judgebox/.style={process, draw=Amber, fill=AmberLight},
  flow/.style={-{Latex[length=2mm]}, draw=Muted, line width=0.5pt},
  lab/.style={font=\scriptsize\bfseries, text=Ink, align=right}
}

\newif\ifresultsready
\resultsreadytrue
\newif\ifaaaitarget
\aaaitargettrue
\title{From Retrieval to Customer Context:\\
Evaluating Frontier-Model Systems for Voice-of-Customer Analysis}
\author{Raviraja G, Viraj Bagal, Prabhath Chellingi}
\affiliations{Enterpret}
\date{}
\begin{document}

\maketitle

\begin{abstract}
Organizations increasingly use frontier language models to analyze customer
feedback, but answer quality also depends on how that feedback is organized and
made available. We define a \emph{customer context graph} as a unified model of
customer and business context. Typed relationships connect customer objects
(feedback, conversations, users, and accounts), operational objects (tickets,
support agents, opportunities, and competitors), and analytical or action
objects (taxonomy concepts, evidence, insights, work items, and outcomes). This
lets an agent investigate not only what customers say, but why, who is affected,
what action followed, who owns it, and whether it was resolved. For this
experiment, the graph is populated from public Cursor feedback; the same
architecture can support any type of feedback source. We
compare Agentic RAG, a Deep Research Agent, and a Customer Context Graph-backed
Agent on the same 9,432 public Cursor feedback records using 30 realistic
product, incident, comparison, and metadata questions.

Without exhaustive ground truth, we jointly score responses on answer
quality (coverage and organization), analytical depth (specificity and
decomposition), and evidence quality (citation support and
traceability), using a comparative rubric calibrated on 28 of the 30
questions. We sample cited records against their
claims and weight the three dimensions equally. Under this aligned
rubric, the Customer Context Graph-backed Agent scores 0.961 overall,
versus 0.710 for the Deep Research Agent and 0.651 for Agentic RAG,
and leads the Deep Research Agent on 27 of 30 paired questions
(sign-test $p<10^{-5}$; strictly best on 26 of 30). Its largest
advantage is analytical depth (0.967 versus 0.642), reflecting more
specific, hierarchically developed findings with quantified themes and
traceable evidence. Because the rubric was calibrated on these
outputs, we report the result as exploratory: the ranking is
internally consistent and survives the two questions withheld from
calibration, but a confirmatory pass on freshly authored questions
scored under the frozen rubric is outstanding and is the validation we
consider decisive. Within this frozen configuration, frontier-model
agents grounded in a persistent customer context graph produce
higher-quality customer-feedback analysis than agents analyzing the
same feedback without that structured context.
\end{abstract}


\section{Introduction}
\label{sec:introduction}

Voice-of-customer (VoC) analysis turns qualitative feedback into a
structured account of customer needs, their relationships, and their
priorities \citep{griffin1993voice}. As more knowledge work is automated,
a fluent answer is not enough: the analysis underneath it must be
correct, reproducible, and traceable, or every downstream product,
engineering, and support decision inherits an unstable understanding of
the customer. Frontier language models make raw feedback easier to
query, but a useful VoC answer still has to combine an exact
denominator, semantic grouping, comparisons across time or versions,
representative quotations, and a defensible statement about what the
data cannot establish. A retrieved passage supports a qualitative
observation but a top-$k$ retrieval result cannot establish corpus
prevalence, and a fluent summary can cite relevant records while
attaching the wrong count or claiming a customer segment absent from
the source data.

Retrieval-augmented generation (RAG) gives a model access to external
evidence \citep{lewis2020rag,karpukhin2020dense}. More capable
query-time systems interleave reasoning with retrieval and tool use
\citep{yao2023react,trivedi2023ircot,asai2024selfrag} or delegate
subtasks to specialist agents. A different architectural choice is to
organize customer context before the question is asked: summarize
feedback, map it to a shared metadata schema, predict record-level
taxonomy labels, retain available user information, and connect these
entities back to each source record. This paper compares those choices
directly on 9{,}432 public feedback records about Cursor---the AI
coding assistant---collected from Reddit developer subreddits, the
Apple App Store review stream, and Trustpilot over the frozen
three-month UTC window 2026-04-16 to 2026-07-15
(\cref{tab:corpus}). Records mix short mobile app reviews,
multi-paragraph Reddit posts, and Trustpilot complaints, and are
collected without knowing which internal product concepts (billing,
agent mode, Auto routing, GLM 5.2 tool calls, Composer) will be
relevant to any analyst question. The comparison deliberately excludes
a naive ``put all reviews in the prompt'' baseline: long-context
models do not use every input position equally \citep{liu2024lost} and
the target corpus exceeds a single request's practical context
capacity.

We ask: how does context architecture affect answer quality, analytical
depth, quantification, evidence quality, and epistemic restraint on
unsupported groupings (RQ1--RQ5), and how much of any quality gap can
stronger query-time planning, retrieval, and evidence review close
(RQ6)?

\paragraph{Contribution.}
Our primary contribution is a controlled three-system comparison
establishing that persistent customer context, not additional query-time
agency, is the load-bearing lever for LLM-based VoC analysis on this
corpus and under an analyst-aligned rubric. Supporting contributions are
(i) the 30-query benchmark and its evaluator-only family invariants;
(ii) a common answer-package contract that retains final prose and
resolvable citation membership; and (iii) a three-part automatic
evaluation of answer quality, analytical depth, and evidence quality
with paired $W/T/L$ as the primary statistic.

\paragraph{Scope.}
The reported comparison is scoped in four ways the reader should carry
through: (1) scores come from a comparative rubric aligned with analyst
preferences after inspecting evaluated answers on these 30 questions
(\cref{sec:data-benchmark}, \cref{sec:limitations}); (2) the comparison
is complete-system rather than component-level; (3) the study uses one
product corpus and one three-month frozen UTC window; and (4) holding
the base analyst model constant does not equalize model capability
across candidates because each system dispatches that model to
structurally different roles (supplementary material).

\section{Related Work}
\label{sec:related}

\paragraph{Voice-of-customer and app-review analysis.}
Classical VoC work separates identifying customer needs, structuring those
needs, and estimating their priority \citep{griffin1993voice}. That
distinction remains useful for LLM-based analysis: extracting representative
statements, organizing them into concepts, and estimating prevalence are
related but non-equivalent tasks. Prior app-review work classifies reviews
into maintenance-relevant categories such as bugs, feature requests, user
experiences, and ratings \citep{panichella2015improve,maalej2016automatic};
a systematic review characterizes app reviews as short, error-prone,
dynamic, and domain-dependent \citep{wang2024mine}; production summarization
uses multi-stage extraction, topic grouping, and human evaluation to
balance freshness, diversity, and accuracy \citep{apple2025reviews}. A
recent VoC benchmark finds that feature-labeled and sentiment-aware RAG
variants improve consistency and coverage of generated product requirements
\citep{fiocco2025voc}. Our focus differs from classification and one-shot
summarization: we compare complete analytical systems over linked questions
that require exhaustive counts, repeated concept use, evidence
reconciliation, and downstream decisions.

\paragraph{RAG, agentic retrieval, and deep-research paradigms.}
RAG combines parametric generation with retrieved non-parametric evidence
\citep{lewis2020rag}; dense passage retrieval provides a standard
candidate-context mechanism \citep{karpukhin2020dense}. Retrieval remains
relevant as context windows grow because models can underuse evidence
placed in the middle of long inputs \citep{liu2024lost}. Agentic and
iterative retrieval interleave reasoning with action
\citep{yao2023react,trivedi2023ircot,asai2024selfrag}. Recent
\emph{deep-research} systems extend these ideas to multi-agent orchestration
with a manager, specialist agents, a verifier, and a finalizer; released
products include OpenAI's Deep Research
\citep{openai2025deepresearch} and Gemini Deep Research
\citep{google2024deepresearch}, and the open literature includes STORM
\citep{shao2024storm} and AutoGen \citep{wu2023autogen}. Our two query-time
candidates instantiate this ladder: \agentic{} is a bounded tool-using
single-analyst workflow, and the \deepresearch{} is a decomposed workflow
with specialist investigations, evidence audit, and revision.

\paragraph{Corpus-level and graph-structured RAG.}
Most RAG benchmarks emphasize local questions whose evidence lies in a few
passages; GlobalQA shows that counting, sorting, and other corpus-level
operations remain difficult for conventional RAG
\citep{luo2025globalrag}. A parallel line of work shifts work from query
time to ingestion time by building a persistent structured index that a
lighter query-time agent then reads. GraphRAG extracts entities and
relations into hierarchical community summaries so that query-focused
summarization is answered from a graph-level summary rather than direct
retrieval \citep{edge2024graphrag}. HippoRAG augments dense retrieval with
Personalized-PageRank traversal \citep{gutierrez2024hipporag}; LightRAG
combines graph indexing with dual-level retrieval
\citep{guo2024lightrag}; RAPTOR builds a recursive tree of clustered
abstractive summaries \citep{sarthi2024raptor}; Structured RAG translates
aggregative questions into formal operations over an ingestion-time
schema \citep{koshorek2025structured}. Our \contextgraph{} shares the
work-at-ingestion premise but differs in what is stored: rather than a
derived-summary graph or an aggregated-chunk index, ingestion persists
typed customer-feedback entities (records, users, sources, themes,
subthemes, sentiment, ratings) with an explicit product taxonomy and
record-level taxonomy predictions that are durable across sessions and
reused across independently executed questions. To our knowledge no prior
benchmark compares a graph-indexed workflow of this kind against agentic
and deep-research workflows on a shared corpus, shared questions, and a
shared answer contract.

\paragraph{Theme reliability and grounded-answer evaluation.}
Statistical topic coherence does not guarantee actionable product
dimensions \citep{nguyen2019siri} and topic-model document assignments can
be unreliable inputs to downstream analysis
\citep{schroeder2025reliability}, so we evaluate theme membership
against adjudicated records rather than trusting coherence alone. RAG
evaluation is multidimensional: RAGAs proposes reference-free metrics
\citep{es2024ragas}, ARES combines automated judges with a human-labeled
set \citep{saadfalcon2024ares}, and ALCE evaluates answer correctness
and citation quality \citep{gao2023alce}. We do not adopt these
evaluators directly because each assumes a single-answer QA setting:
RAGAs and ALCE score a system answer against a query with a known
correct answer or a small set of ground-truth passages; ARES trains a
judge on a human-labeled QA dataset. Corpus-level analytical answers
that combine typed classification counts, aggregate drill-down
citations, per-theme quantification, and refusal on unsupported
groupings do not fit that mold: there is no single ``correct'' answer
against which to measure precision or faithfulness. We instead
evaluate paired candidate answers and audit their cited records
directly, following the attribution-to-identified-source protocol of
\citet{rashkin2023ais} and the atomic factuality intuition of
\citet{min2023factscore}. LLM judges scale open-ended evaluation but
exhibit position, verbosity, and self-enhancement biases
\citep{zheng2023judge}, so we use query-specific anonymized answer
order, separate comparative dimensions, and deterministic citation
sampling. Public reviews are also a self-selected view of customer
experience and review solicitation changes extremity and
representativeness \citep{karaman2021representativeness}; our
prevalence measurements therefore describe the frozen corpus, not the
full Cursor customer population.

\section{Compared Systems}
\label{sec:systems}

\begin{figure*}[t]
\centering
\resizebox{0.99\textwidth}{!}{
\begin{tikzpicture}[
  x=1mm,y=1mm,
  stage/.style={font=\scriptsize\bfseries,text=Muted,align=center},
  box/.style={process,minimum height=13mm},
  answer/.style={process,text width=23mm,minimum height=13mm},
  junction/.style={circle,fill=Muted,inner sep=1.2pt}
]
  \node[stage] at (13,39) {Shared input};
  \node[stage] at (48,39) {System-specific preparation};
  \node[stage] at (90,39) {Question execution};
  \node[stage] at (130,39) {Answer package};
  \node[stage] at (164,39) {Common evaluation};
  \node[stage] at (195,39) {Output};

  \node[box,text width=25mm] (corpus) at (13,8) {Neutral parser\\9,432 feedback\\records + metadata};
  \node[junction] (fork) at (28,8) {};
  \draw[flow] (corpus.east) -- (fork);

  \node[managedbox,text width=29mm,minimum height=13mm] (a-prep) at (48,28)
    {\textbf{Agentic RAG}\\vector index\\metadata store};
  \node[managedbox,text width=35mm,minimum height=13mm] (a-run) at (90,28)
    {Single analyst agent\\plans + retrieves\\aggregates + synthesizes};
  \node[answer] (a-out) at (130,28) {Cited answer\\record evidence};

  \node[judgebox,text width=29mm,minimum height=13mm] (d-prep) at (48,8)
    {\textbf{Deep Research Agent}\\vector index\\exact-analysis tools};
  \node[judgebox,text width=35mm,minimum height=13mm] (d-run) at (90,8)
    {Manager + specialists\\evidence ledger, verifier, revision};
  \node[answer] (d-out) at (130,8) {Cited answer\\record + ledger evidence};

  \node[structurebox,text width=29mm,minimum height=13mm] (e-prep) at (48,-12)
    {\textbf{Context Graph Agent}\\summaries + predictions\\metadata + user links};
  \node[structurebox,text width=35mm,minimum height=13mm] (e-run) at (90,-12)
    {Agent groups + counts\\compares + drills down};
  \node[answer] (e-out) at (130,-12) {Cited answer\\aggregate + record evidence};

  \node[junction] (join) at (146,8) {};
  \node[judgebox,text width=27mm,minimum height=25mm] (eval) at (164,8)
    {Paired evaluator\\answer quality\\analytical depth\\evidence quality};
  \node[process,text width=23mm,minimum height=17mm] (result) at (195,8)
    {Per-query scores\\+ aggregate comparison};

  \draw[flow] (fork) |- (a-prep.west);
  \draw[flow] (fork) -- (d-prep.west);
  \draw[flow] (fork) |- (e-prep.west);
  \draw[flow] (a-prep) -- (a-run); \draw[flow] (a-run) -- (a-out);
  \draw[flow] (d-prep) -- (d-run); \draw[flow] (d-run) -- (d-out);
  \draw[flow] (e-prep) -- (e-run); \draw[flow] (e-run) -- (e-out);
  \draw[flow] (a-out.east) -| (join);
  \draw[flow] (d-out.east) -- (join);
  \draw[flow] (e-out.east) -| (join);
  \draw[flow] (join) -- (eval); \draw[flow] (eval) -- (result);
\end{tikzpicture}}
\caption{End-to-end benchmark architecture. A neutral parsing layer supplies
the same feedback text and source metadata to all candidates. Each system keeps
its native preparation and question-execution architecture, but all outputs are
converted to a common answer package and scored by the same paired evaluator.}
\label{fig:system-comparison}
\end{figure*}
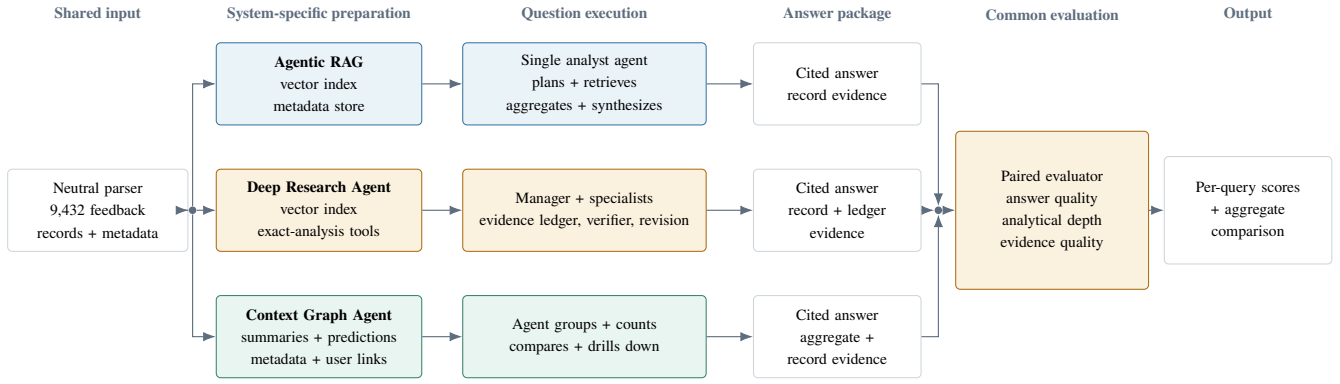

We compare three concrete implementations
(\cref{fig:system-comparison}); candidate names refer to these
implementations, not to all systems in an architectural family. Each
receives the same query text and evidence from the same frozen corpus.
We use one running question, \texttt{q013} (\emph{``Are there signs of
a recently increasing reliability issue in Cursor?''}), to illustrate
what each system does differently; it requires an aligned time
comparison, an explicit denominator, per-cluster counts, and a
defensible refusal when a comparable baseline is not available.

\paragraph{\agentic{}} is one customer-feedback analyst built with an
agent SDK: bounded tool use, semantic search, record inspection, exact
metadata aggregation, segment comparison, and cited synthesis. On
\texttt{q013} it runs its own 30-day preceding/recent comparison with
a keyword reliability proxy (96/557 vs 851/8{,}652 records) and labels
the trend low-confidence. Any semantic groups it creates belong to
that run and do not persist across questions.

\paragraph{\deepresearch{}} decomposes each question into typed
evidence requirements: a manager delegates to qualitative,
quantitative, and incident specialists that search, inspect records,
aggregate exactly, and---when enabled---classify every record in a
filter-derived denominator (cached with a record-level ledger and
content hash); a verifier audits claims, counts, coverage, and
citations before a finalizer revises the answer. On \texttt{q013} it
constructs matched rolling 14-day windows (3{,}720 vs 3{,}717 records)
so all three sources are present in both and localizes a GLM 5.2
tool-call cluster to two specific dates. Query-specific assignments
are not exposed to other questions as a shared ontology or durable
label store.

\paragraph{\contextgraph{}} receives normalized records after they are
processed into a persistent customer context graph. Before benchmark
questions are asked the system summarizes feedback, induces a product
taxonomy, stores record-level taxonomy predictions, maps source fields
to a shared metadata schema, retains available user information, and
stores relationships among these entities. The query-time agent
traverses those relationships to group, count, compare, and cite
feedback. On \texttt{q013} it reads 86 reliability-adjacent themes
already tagged in the graph, aggregates conversation counts per theme,
and declines a trend claim because the preceding calendar window is 24
Trustpilot-only conversations against 9{,}408 recent mixed-source
conversations.

\begin{figure*}[t]
\centering
\resizebox{0.99\textwidth}{!}{
\begin{tikzpicture}[
  x=1mm,y=1mm,
  title/.style={font=\scriptsize\bfseries,align=left},
  component/.style={process,text width=27mm,minimum height=13mm},
  widecomponent/.style={process,text width=34mm,minimum height=13mm},
  panel/.style={draw=Rule,rounded corners=2pt,inner sep=5mm}
]

  \node[title,text=Blue] (atitle) at (0,45) {Agentic RAG: one tool-using analyst};
  \node[component] (aq) at (16,30) {User question};
  \node[managedbox,text width=31mm,minimum height=13mm] (agent) at (52,30)
    {Analyst agent\\plans retrieval\\and revises};
  \node[managedbox,text width=39mm,minimum height=13mm] (atools) at (96,30)
    {Semantic search + metadata\\exact aggregates + record inspection};
  \node[managedbox,text width=31mm,minimum height=13mm] (asynth) at (137,30)
    {Transient themes\\counts + synthesis};
  \node[component] (aa) at (173,30) {Cited answer};
  \draw[flow] (aq) -- (agent); \draw[flow] (agent) -- (atools);
  \draw[flow] (atools) -- (asynth); \draw[flow] (asynth) -- (aa);
  \node[panel,fit=(atitle)(aq)(aa)] {};

  \node[title,text=Amber] (dtitle) at (0,4) {Deep Research Agent: managed, parallel analysis with verification};
  \node[component] (dq) at (16,-11) {User question};
  \node[judgebox,text width=31mm,minimum height=13mm] (manager) at (52,-11)
    {Manager\\decomposes\\evidence needs};
  \node[judgebox,text width=39mm,minimum height=13mm] (specialists) at (96,-11)
    {Qualitative + quantitative\\incident specialists\\shared retrieval tools};
  \node[judgebox,text width=31mm,minimum height=13mm] (verify) at (137,-11)
    {Evidence ledger\\verifier + finalizer};
  \node[component] (da) at (173,-11) {Cited answer};
  \draw[flow] (dq) -- (manager); \draw[flow] (manager) -- (specialists);
  \draw[flow] (specialists) -- (verify); \draw[flow] (verify) -- (da);
  \node[panel,fit=(dtitle)(dq)(da)] {};

  \node[title,text=Green] (etitle) at (0,-37) {Customer Context Graph-backed Agent: stored feedback context queried at run time};
  \node[component] (records) at (16,-52) {Normalized records\\+ metadata};
  \node[structurebox,text width=31mm,minimum height=13mm] (process) at (52,-52)
    {Feedback summaries\\taxonomy predictions\\metadata mapping};
  \node[structurebox,text width=39mm,minimum height=13mm] (context) at (96,-52)
    {Stored customer context\\records + summaries\\predictions + metadata\\user information + links};
  \draw[flow] (records) -- (process); \draw[flow] (process) -- (context);

  \node[component] (eq) at (16,-73) {User question};
  \node[structurebox,text width=31mm,minimum height=13mm] (eagent) at (52,-73)
    {Context Graph Agent\\maps question to\\product concepts};
  \node[structurebox,text width=39mm,minimum height=13mm] (etools) at (96,-73)
    {Graph queries + aggregation\\conversation drill-down};
  \node[structurebox,text width=31mm,minimum height=13mm] (esynth) at (137,-73)
    {Specific themes\\quantification + synthesis};
  \node[component] (ea) at (173,-73) {Cited answer};
  \draw[flow] (eq) -- (eagent); \draw[flow] (eagent) -- (etools);
  \draw[flow] (context) -- (etools); \draw[flow] (etools) -- (esynth);
  \draw[flow] (esynth) -- (ea);
  \node[panel,fit=(etitle)(records)(eq)(ea)] {};
\end{tikzpicture}}
\caption{Component architecture of the three evaluated candidates. Agentic RAG
uses one analyst over retrieval and exact-analysis tools. The Deep Research Agent
adds decomposition, specialist roles, an evidence ledger, verification, and
revision over that retrieval layer. The context graph system first connects
feedback, summaries, taxonomy predictions, metadata, and available user
information; its agent then computes aggregates and drills down to source
records when producing an answer.}
\label{fig:semantic-lifecycle}
\end{figure*}
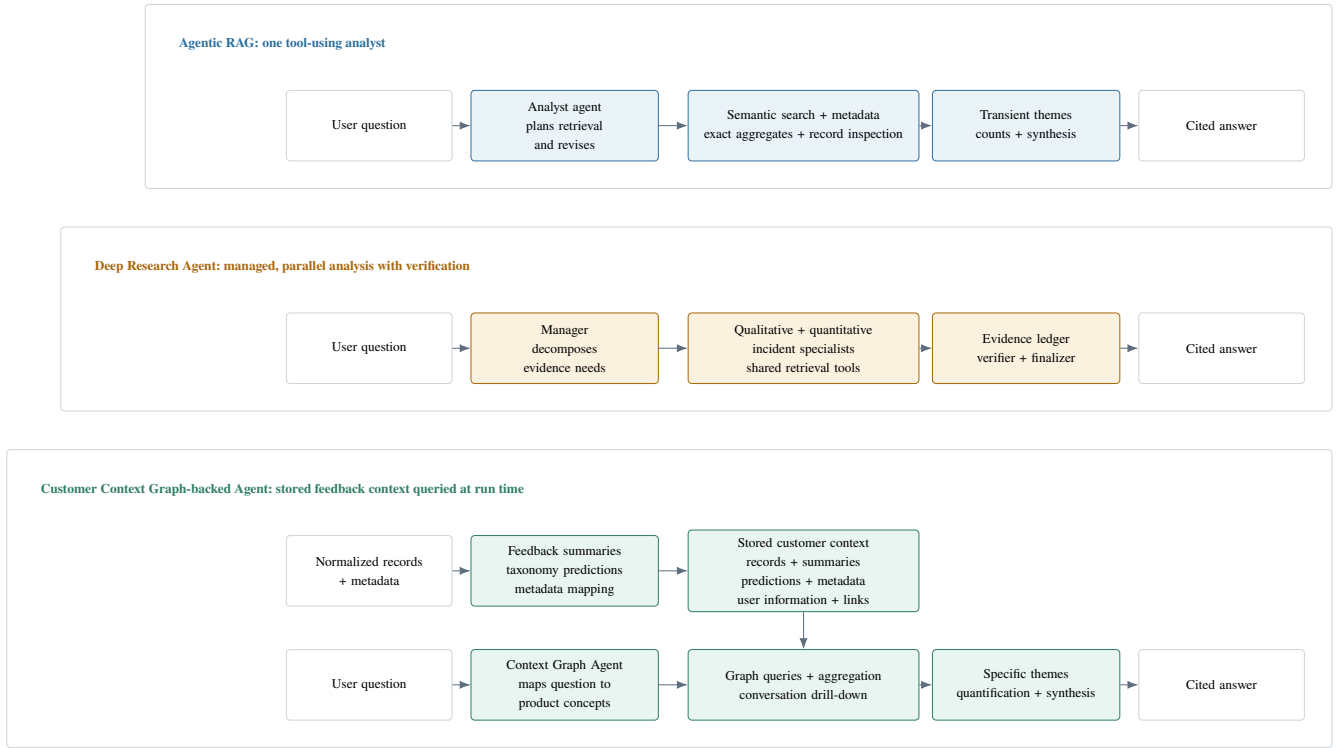

\paragraph{Experimental controls and complete-system treatment.}
\Cref{tab:controls} states what was held constant, what was allowed to
differ, and what the experiment can therefore claim; every run records
model and tool versions, prompts, token use, tool calls, failures,
corpus hash, and configuration hash. Failed or incomplete runs remain
in the denominator with no credit, so a non-completion cannot silently
drop out of the average. Freeze order and rubric calibration are
detailed in \cref{sec:data-benchmark}.

\begin{table}[t]
\centering
\caption{Experimental controls for the three-system comparison.}
\label{tab:controls}
\small
\begin{tabular}{L{0.22\columnwidth}L{0.68\columnwidth}}
\toprule
\textbf{Held constant} &
Frozen 9{,}432-record corpus + SHA-256; identical query text and
as-of date; base analyst model, both judges, retrieval reranker;
answer contract; retry policy; alias-shuffled evaluator presentation. \\
\textbf{Allowed to differ (the treatment)} &
Whether the system reconstructs product concepts per query or reads
pre-computed summaries, taxonomy predictions, metadata, user
information, and graph relationships stored at ingestion; the
orchestration budget the analyst model is dispatched into. \\
\textbf{Can therefore claim} &
The complete-system effect of persistent customer context on this
corpus, question set, and calibrated rubric; not the isolated causal
effect of any component in isolation. \\
\bottomrule
\end{tabular}
\end{table}

\section{Dataset and Benchmark}
\label{sec:data-benchmark}

\paragraph{Corpus.}
The evaluated corpus contains 9{,}432 public Cursor feedback records from
the frozen three-month UTC window 2026-04-16 through 2026-07-15
(\cref{tab:corpus}). Collection, normalization, and deduplication are
frozen before evaluation; each record preserves a stable identifier,
source, timestamp, feedback text, source URL, rating when available, and
public metadata. Missing account or commercial attributes remain missing.
All reported prevalence describes this frozen corpus rather than the full
Cursor user base.

\begin{table}[t]
\centering
\caption{Frozen Cursor corpus used by all three systems.}
\label{tab:corpus}
\small
\begin{tabular}{L{0.42\columnwidth}rr}
\toprule
Statistic & Value & Coverage \\
\midrule
Normalized feedback records & 9{,}432 & 100\% \\
Collection window & \multicolumn{2}{c}{2026-04-16--2026-07-15 UTC} \\
Public feedback sources & 3 & 100\% \\
Records with rating & 171 & 1.81\% \\
Records with normalized version & 5 & 0.05\% \\
Records with public country & 118 & 1.25\% \\
Account/commercial cohort ID & 0 & 0\% \\
Corpus SHA-256 & \multicolumn{2}{c}{\texttt{d62fa842\ldots f23a62f0}} \\
\bottomrule
\end{tabular}
\end{table}

\paragraph{Information parity and denominator contract.}
All systems receive feedback originating from the same frozen
normalized records; source URLs and public metadata are preserved
consistently, and the two query-time systems do not receive
taxonomies, summaries, clusters, or other labels generated by the
context graph system. The context graph system's ingestion
additionally groups the 9{,}432 records into 5{,}035 customer
\emph{conversations} by threading records that share a parent post,
comment chain, review reply chain, or platform-linked exchange within
the same source and author. To keep quantified claims comparable at
judge time we require a denominator contract: every quantified claim
must state the total it is divided by and cite the record set that
produced that total. Baselines use the record count (they do not
receive conversation-adjacency edges); the context graph candidate may
report at conversation grain but must state the conversation total and
the constituent record count. The comparative evidence-strength score
rewards explicit and cited denominators over absolute magnitude, so
180 conversations with denominator and drill-down cannot outscore 892
records with equivalent evidence structure by being smaller. After a
pilot exposed a 123-record snapshot mismatch we refreshed the neutral
export and re-ran all three candidates on all 30 queries against the
repaired corpus and its SHA-256 hash. Per-run identifiers and the
canonical evaluation-pass identifier are recorded in the answer
packages and in the source snapshot macros; the build
gate refuses to update either without matching \texttt{run\_metadata}
across all 90 cells. Query contracts and evaluator prompts are hidden
from every system; each query executes in a fresh conversation.

\paragraph{Query benchmark.}
The benchmark contains 30 product-adapted questions arranged into seven
analysis families (executive VoC, product area, reliability and incidents,
pricing and packaging, source and metadata, competition and retention, and
cross-functional action; sizes $\{5,5,5,5,5,4,1\}$) and fixed before
candidate execution, spanning four depth levels (\cref{tab:query-depth});
the same question text and as-of date are sent byte-for-byte to every
candidate. Hidden family invariants test exact-value reuse, semantic
stability, segment-level summaries, trend direction, sampling restraint,
and whether recommended actions trace to previously measured evidence.
Each query is a small contract: it names the operations the answer
must perform, the fields it may draw on, the expected output
elements, count semantics, citation requirements, evaluation scope,
gold source, and known limitations. The benchmark has no independent
exhaustive semantic labels. For semantic claims the LLM judge scores
only records actually supplied by each candidate, so we can estimate
evidence validity on observed assignments but cannot identify false
negatives; semantic recall and F1 are therefore not reported.

\begin{table}[t]
\centering
\caption{Benchmark depth. Counts are fixed by the query manifest.}
\label{tab:query-depth}
\small
\begin{tabular}{cL{0.58\columnwidth}r}
\toprule
Depth & Required reasoning & $n$ \\
\midrule
1 & Exact lookup, metadata coverage, capability checks & 11 \\
2 & Focused semantic classification or evidence synthesis & 10 \\
3 & Comparisons across source, time, or cohort & 3 \\
4 & Multi-stage VoC, strategy, incident, and workflow decisions & 6 \\
\midrule
 & \textbf{Total} & \textbf{30} \\
\bottomrule
\end{tabular}
\end{table}

\paragraph{Freeze order, calibration, and held-out confirmation.}
We freeze, in order, (1) research questions and system definitions;
(2) query text and answer contracts; (3) collection, normalization,
and deduplication; (4) candidate prompts, tools, models, and budgets;
(5) an initial judge rubric v1; (6) a first evaluated run against v1;
(7) a bounded calibration pass in which the judge prompts were
adjusted after human inspection of v1 outputs on the first 27
questions, producing rubric v2; (8) a re-evaluation against v2; and
(9) a single further narrowing of the competitor-gap criterion,
prompted by inspecting the q029 answers, after which only q029 was
rescored. Steps (1)--(9) produced the scores reported in
\cref{sec:results}, so 28 of the 30 questions informed the rubric that
scores them. We verified that v2 applied to q001--q027 yields
identical per-cell scores to the earlier calibrated-on-q001--q027
snapshot, confirming that steps 7--8 did not retroactively move the
rubric on the calibration set. Because v2 was reached while inspecting
q001--q027, questions q028 and q030 were scored under it without
further change and provide a small internal stability check
(\cref{sec:results}); q029 is excluded because its rubric was narrowed
after its answers were inspected. We deliberately do not call this a
confirmatory set: two same-session, same-family questions withheld
from calibration are not equivalent to freshly authored questions, and
the reported margins should be read as exploratory. The step 6--8
calibration is a deviation from a purely sequential freeze order,
treated as the principal threat in \cref{sec:limitations}.

\section{Automatic Evaluation}
\label{sec:evaluation}

\begin{figure}[t]
\centering
\resizebox{0.98\columnwidth}{!}{
\begin{tikzpicture}[node distance=4mm and 4mm]
  \node[process, minimum width=31mm] (answers) {Three anonymized answers};
  \node[judgebox, below left=6mm and -1mm of answers, text width=24mm] (compare)
    {Comparative answer judge};
  \node[managedbox, below right=6mm and -1mm of answers, text width=24mm] (sample)
    {Up to 15 records per citation};
  \node[judgebox, below=of sample, text width=24mm] (evidence)
    {Claim--record relevance judge};
  \node[structurebox, below left=8mm and -1mm of evidence, text width=25mm] (answerdims)
    {Answer quality + analytical depth};
  \node[structurebox, below right=8mm and -1mm of evidence, text width=25mm] (evidencedim)
    {Evidence quality + audit coverage};
  \node[structurebox, below=8mm of answerdims, xshift=15mm, text width=34mm] (report)
    {Overall mean + reasons + paired results};

  \foreach \x/\y in {answers/compare,answers/sample,sample/evidence,
    compare/answerdims,evidence/evidencedim,answerdims/report,evidencedim/report}
    \draw[flow] (\x) -- (\y);
\end{tikzpicture}}
\caption{The simple protocol compares anonymized answers and audits every
resolvable citation group separately.}
\label{fig:evaluation-pipeline}
\end{figure}
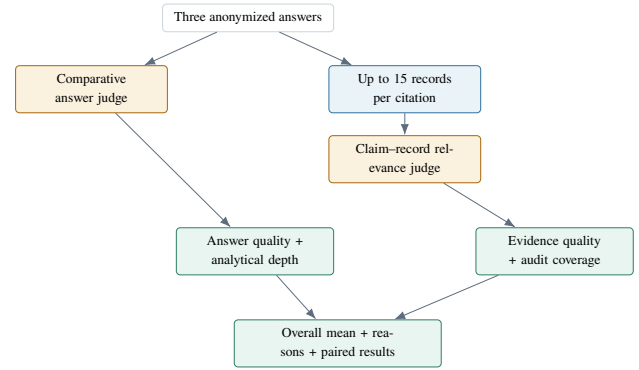

Open-ended customer-feedback questions have no exhaustive reference
answer. We therefore compare all three answers for each query and
separately audit whether cited records support the claims made
(\cref{fig:evaluation-pipeline}). The query is the unit of evaluation,
all candidates are judged together under anonymized aliases with
deterministically shuffled order, and every report retains
dimension-level rationales and citation audits. The three reported
dimensions and their scoring conventions are summarized here; full
protocol details, macro/micro-average conventions, and citation-audit
provenance appear in the supplementary material.

\paragraph{Answer quality and analytical depth.}
The comparative judge places candidates into best-to-worst tiers on each
dimension. Answer quality rewards explicit question coverage, organization
for the intended reader, useful quantification, and visible separation of
evidence, interpretation, and caveats. Analytical depth rewards a hierarchy
from decision-relevant conclusion to bounded themes, drivers, subthemes,
counterevidence, and synthesis; a one-problem test penalizes umbrella
labels that merge independently actionable issues. With three distinct
tiers, scores are $\{1.0, 0.75, 0.5\}$; two-tier verdicts map to
$\{1.0, 0.5\}$ with the middle tier reserved for three-tier verdicts. Both
dimensions therefore fall in $\{0.5, 0.75, 1.0\}$ for completed answers.

\paragraph{Evidence quality.}
Evidence quality $E_q$ is the harmonic mean of sampled citation
precision $P_q$ and a comparative evidence-strength tier score $S_q$:
\begin{equation}
  P_q = \frac{N_{\mathrm{rel}} + 0.5\,N_{\mathrm{part}}}{N_q},
  \qquad
  E_q = \frac{2P_qS_q}{P_q+S_q}.
\end{equation}
For each record citation and each aggregate citation whose members are
exported, the evaluator deterministically samples up to 15 frozen records
and a claim--record judge labels each pair \emph{relevant}, \emph{partially
relevant}, or \emph{irrelevant}, following the attribution-to-identified-
source protocol of \citet{rashkin2023ais} and the atomic factuality
intuition of \citet{min2023factscore}. $S_q$ rewards claim-level support,
explicit denominators, aggregate drill-down links, and representative
record citations. An answer without citations receives zero. Aggregate
citations with valid drill-down metadata but without exported members are
counted as \emph{traceable}, not member-sampled; unresolved citations are
reported separately. Evidence quality therefore evaluates sampled
correctness and comparative evidence design; it does not imply that every
record behind every aggregate was independently verified.

\paragraph{Overall score and aggregation.}
The overall score combines the three dimensions equally,
$\metric{Overall}_q = (Q_q + D_q + E_q)/3$, and we macro-average query-level
scores across the 30 queries so each query weighs equally in the system
total. Paired win, tie, and loss counts follow the paired-evaluation
formulation of \citet{peyrard2021paired} with a prespecified 0.01
absolute-difference tie tolerance; paired $W/T/L$ is the primary
system-comparison statistic and $\overline{\Delta}$ is descriptive.
Each (query, candidate) cell is a \emph{single} run scored by one judge
pass; we do not average over reruns, so run-to-run variance is
unmeasured and bootstrap intervals quantify variation across queries
rather than across reruns. LLM judges can also exhibit position,
verbosity, and
self-enhancement biases \citep{zheng2023judge}; the alias-shuffling and
structured-output requirements above reduce but do not eliminate those
risks.

\ifaaaitarget
\section{Results}
\label{sec:results}
\label{sec:analysis}

All three systems completed all 30 questions, yielding 90 dimension-level
reports under protocol \texttt{simple-comparison-v30}. The query is the
scoring unit; we macro-average query-level scores and compare systems on
paired questions with a 0.01 tie tolerance. Category means and per-depth
slices are descriptive because families contain between 1 and 5 queries,
and semantic recall is not reported because no independent exhaustive set
identifies every record relevant to an open-ended feedback theme.

\begin{table*}[t]
\centering
\caption{Macro-averaged automatic evaluation across 30 paired queries.
Overall combines the three dimensions equally. Per-dimension scores for
answer quality and analytical depth are drawn from a discrete
comparative-tier scale on $[0.5,1]$ for completed answers with 0.25
spacing (see \cref{sec:evaluation}); evidence quality is a harmonic mean
of sampled citation precision and comparative evidence-strength score in
$[0,1]$. Because the scale is discrete and the primary comparison is
paired, means are descriptive; the primary comparison statistic is
paired $W/T/L$ (\cref{tab:paired}). Best values in bold; bracketed
overall values are bootstrap 95\% CIs over 30 queries ($B=5{,}000$,
seed 42).}
\label{tab:overall}
\small
\begin{tabular}{lccccc}
\toprule
System & Overall (95\% CI) & Answer quality & Analytical depth & Evidence quality & Queries \\
\midrule
Agentic RAG & 0.651 [0.613, 0.692] & 0.625 & 0.642 & 0.686 & 30/30 \\
Deep Research Agent & 0.710 [0.668, 0.752] & 0.683 & 0.642 & 0.806 & 30/30 \\
Context Graph Agent & \textbf{0.961} [0.936, 0.983] & \textbf{0.958} & \textbf{0.967} & \textbf{0.958} & 30/30 \\
\bottomrule
\end{tabular}
\end{table*}

The \contextgraph{} obtains the highest score on every dimension
(\cref{tab:overall}), with overall 0.961 (0.251 above the \deepresearch{},
0.310 above \agentic{}); its largest gap is analytical depth (0.967 vs
0.642). The \deepresearch{} improves overall over \agentic{} by 0.059,
driven primarily by evidence quality (0.806 vs 0.686).

\begin{table*}[t]
\centering
\caption{Paired query-level comparisons with a 0.01 tie tolerance. $W/T/L$
counts are from the perspective of the first system; $\overline{\Delta}$ is
its mean score minus the second system's score, with a bootstrap 95\%
confidence interval over 30 queries ($B=5{,}000$ resamples, seed 42); $p$
reports a two-sided sign test on non-tied paired outcomes. Because the
completed-answer score scale is discrete with 0.25 spacing (see
\cref{tab:overall} and \cref{sec:evaluation}), the paired $W/T/L$ result is
the primary comparison statistic and $\overline{\Delta}$ is descriptive.}
\label{tab:paired}
\small
\begin{tabular}{llrrrlr}
\toprule
Comparison & Metric & Wins & Ties & Losses & $\overline{\Delta}$ (95\% CI) & Sign-test $p$ \\
\midrule
\multirow{4}{*}{Context graph $-$ Deep}
 & Overall & 27 & 1 & 2 & +0.251 [+0.193, +0.305] & $1.6\!\times\!10^{-6}$ \\
 & Answer quality & 25 & 1 & 4 & +0.275 & $1.0\!\times\!10^{-4}$ \\
 & Analytical depth & 28 & 0 & 2 & +0.325 & $8.7\!\times\!10^{-7}$ \\
 & Evidence quality & 27 & 2 & 1 & +0.153 & $2.2\!\times\!10^{-7}$ \\
\midrule
\multirow{4}{*}{Context graph $-$ Agentic}
 & Overall & 29 & 0 & 1 & +0.310 [+0.263, +0.354] & $5.8\!\times\!10^{-8}$ \\
 & Answer quality & 28 & 1 & 1 & +0.333 & $1.1\!\times\!10^{-7}$ \\
 & Analytical depth & 28 & 0 & 2 & +0.325 & $8.7\!\times\!10^{-7}$ \\
 & Evidence quality & 28 & 1 & 1 & +0.272 & $1.1\!\times\!10^{-7}$ \\
\midrule
\multirow{4}{*}{Deep $-$ Agentic}
 & Overall & 17 & 0 & 13 & +0.059 [$-$0.018, +0.135] & 0.585 \\
 & Answer quality & 17 & 0 & 13 & +0.058 & 0.585 \\
 & Analytical depth & 15 & 0 & 15 & +0.000 & 1.000 \\
 & Evidence quality & 26 & 0 & 4 & +0.120 & $6.0\!\times\!10^{-5}$ \\
\bottomrule
\end{tabular}
\end{table*}

\paragraph{Paired outcomes.}
The \contextgraph{} leads the \deepresearch{} on 27 of 30 overall queries,
ties once, and loses twice (\cref{tab:paired}; sign test on non-tied
outcomes, $p=1.6\!\times\!10^{-6}$; bootstrap 95\% CI on
$\overline{\Delta}_{\text{overall}}$ $[+0.193, +0.305]$). It leads
analytical depth on 28 of 30 ($p=8.7\!\times\!10^{-7}$) and leads
\agentic{} on 29 of 30 overall ($p=5.8\!\times\!10^{-8}$;
$\overline{\Delta}_{\text{overall}}=+0.310$, 95\% CI
$[+0.263,+0.354]$). The \contextgraph{} is strictly best on 26 of 30
individual queries; the four exceptions are \texttt{q003} (tied with
\deepresearch{}), \texttt{q006} and \texttt{q013} (lost to
\deepresearch{} on autocomplete overview and matched-window reliability),
and \texttt{q022} (``What are people praising in the App Store?'', which
\agentic{} wins 0.887 vs 0.833 by combining tier-1 answer quality and
depth with a lower evidence-quality score). The secondary
\deepresearch{}-vs-\agentic{} comparison is less monotonic: \deepresearch{}
wins 17 overall and loses 13 (sign test $p=0.585$;
$\overline{\Delta}_{\text{overall}}=+0.059$, 95\% CI $[-0.018,+0.135]$),
its evidence-quality advantage is real (26/4 wins, $p=6.0\!\times\!
10^{-5}$), but analytical-depth outcomes split 15--15 and average to
zero.

\begin{table*}[t]
\centering
\caption{Citation-audit coverage. ``Member-audited'' citations have at least one
frozen record judged against the nearby claim. Traceable aggregates retain
drill-down metadata but were not member-sampled.}
\label{tab:evidence}
\small
\begin{tabular}{lrrrrr}
\toprule
System & Cited & Member-audited & Traceable aggregate & Unresolved & Records sampled \\
\midrule
Agentic RAG & 712 & 708 & 0 & 4 & 708 \\
Deep Research Agent & 594 & 594 & 0 & 0 & 594 \\
Context Graph Agent & 913 & 127 & 786 & 0 & 127 \\
\bottomrule
\end{tabular}
\end{table*}

\paragraph{Citation-audit coverage.}
\agentic{} emits 712 citations and \deepresearch{} emits 594; nearly all
resolve directly to member records (708 and 594 sampled records
respectively). The \contextgraph{} emits 913 citations of which 127 are
member-audited and 786 are aggregate citations with valid drill-down
metadata but without exported members
(\cref{tab:evidence}). No context-graph citation is unresolved. Its high
evidence-quality score therefore reflects perfect sampled precision on
exported member evidence together with the comparative judge's preference
for claim-level quantified evidence architecture; it must not be read as
independent validation of all records behind the 786 traceable
aggregates.

\paragraph{Partial internal rubric-stability check.}
The v2 rubric was reached while inspecting q001--q027 and applied
unchanged to q028 and q030 (q029 is excluded: its rubric was narrowed
after its answers were inspected). On that $n=2$ slice the
\contextgraph{} wins both queries against both baselines
($\overline{\Delta}_{\text{overall}}$ $=+0.302$ versus the
\deepresearch{}, $+0.404$ versus \agentic{}), consistent with the
calibration set. This is a weak internal check, not confirmation:
$n=2$ carries almost no inferential power, and q028 and q030 were
authored in the same session and families as the calibration queries,
so a rubric shaped on their siblings can carry correlated bias into
them. Withholding a question from calibration is not equivalent to
drawing a fresh one. The decisive test---new queries authored after the
rubric was frozen, scored without further rubric change---is not
performed here and remains the primary outstanding validation
(\cref{sec:limitations}).

\paragraph{Findings across RQs and interpretation.}
\label{sec:rq-summary}
Under the aligned rubric, the \contextgraph{} leads on every dimension
in \cref{tab:overall} (RQ1), and its lead is stable across all four
benchmark depth levels (RQ2, supplementary material). Quantification
splits the systems by grain: baselines report exclusive-classification
counts across the full record set, while the \contextgraph{} reports
theme counts at conversation grain with per-theme drill-down (RQ3).
The citation audit shows that most of the context-graph
evidence-quality advantage comes from aggregate drill-down citations,
not from higher sampled precision (RQ4). On epistemic restraint the
strongest in-run signal is \texttt{q013}, where the \contextgraph{}
refuses a period-over-period trend claim; a pre-registered refusal
battery is future work (RQ5). And on structure versus agency, the
\deepresearch{}'s decomposition, verifier, and revision loop lift
evidence quality over \agentic{} but leave overall and depth outcomes
indistinguishable at $n=30$ (RQ6).

The dimension pattern suggests a straightforward mechanism.
Ingestion-time processing makes stored summaries, taxonomy predictions,
metadata, and relationships \emph{durable}: at query time the agent
traverses those relationships to bounded themes with aggregate-citation
quantification rather than reconstructing product concepts each
question. Even so, the result is not universal dominance. The
\deepresearch{} wins the two questions that reward bespoke matched
comparisons (\texttt{q006}, \texttt{q013}), and \agentic{} wins the
App-Store-praise question (\texttt{q022}). Query-specific computation
can therefore beat stored concepts when a bespoke comparison matters
more than concept reuse. Representative response excerpts (including
the \texttt{q013} reconciliation), per-depth and per-family means, and
the extended mechanism analysis are in the supplementary material.
\else
\section{Analysis}
\label{sec:analysis}

The benchmark question is the scoring unit. All three systems answered the same
30 questions, yielding 90 candidate reports. We macro-average query-level
scores, then compare systems on paired questions with a 0.01 tie tolerance. The
primary comparison is \contextgraph{} versus \deepresearch{}: whether the strongest
query-time reconstruction workflow closes the gap to stored customer context.
The secondary comparison is \deepresearch{} versus \agentic{}, measuring the
gain from decomposition, broader classification, specialist analysis, evidence
review, and revision. These are complete-system comparisons rather than
component-level causal ablations.

Category means are descriptive because the seven families contain between one
and five questions. We do not report semantic recall: no independent exhaustive
set identifies every record relevant to an open-ended feedback theme. Citation
counts and sampled-record coverage accompany the quality scores so that a high
evidence score is not mistaken for exhaustive verification of every aggregate.

\section{Results}
\label{sec:results}

All systems completed all 30 questions. The final artifact contains 90
dimension-level reports under protocol \texttt{simple-comparison-v30}.

\subsection{Overall Comparison}

\begin{table*}[t]
\centering
\caption{Macro-averaged automatic evaluation across 30 paired queries.
Overall combines the three dimensions equally. Per-dimension scores for
answer quality and analytical depth are drawn from a discrete
comparative-tier scale on $[0.5,1]$ for completed answers with 0.25
spacing (see \cref{sec:evaluation}); evidence quality is a harmonic mean
of sampled citation precision and comparative evidence-strength score in
$[0,1]$. Because the scale is discrete and the primary comparison is
paired, means are descriptive; the primary comparison statistic is
paired $W/T/L$ (\cref{tab:paired}). Best values in bold; bracketed
overall values are bootstrap 95\% CIs over 30 queries ($B=5{,}000$,
seed 42).}
\label{tab:overall}
\small
\begin{tabular}{lccccc}
\toprule
System & Overall (95\% CI) & Answer quality & Analytical depth & Evidence quality & Queries \\
\midrule
Agentic RAG & 0.651 [0.613, 0.692] & 0.625 & 0.642 & 0.686 & 30/30 \\
Deep Research Agent & 0.710 [0.668, 0.752] & 0.683 & 0.642 & 0.806 & 30/30 \\
Context Graph Agent & \textbf{0.961} [0.936, 0.983] & \textbf{0.958} & \textbf{0.967} & \textbf{0.958} & 30/30 \\
\bottomrule
\end{tabular}
\end{table*}

The \contextgraph{} obtains the highest score on every dimension
(\cref{tab:overall}). Its overall score is 0.961, 0.251 above the Deep Research
Agent and 0.310 above Agentic RAG. The largest context-graph--deep-research gap is analytical
depth (0.967 versus 0.642), followed by answer quality (0.958 versus 0.683).
The Deep Research Agent improves the overall score over Agentic RAG by 0.059,
driven primarily by its evidence-quality score (0.806 versus 0.686).

\subsection{Paired Outcomes}

\begin{table*}[t]
\centering
\caption{Paired query-level comparisons with a 0.01 tie tolerance. $W/T/L$
counts are from the perspective of the first system; $\overline{\Delta}$ is
its mean score minus the second system's score, with a bootstrap 95\%
confidence interval over 30 queries ($B=5{,}000$ resamples, seed 42); $p$
reports a two-sided sign test on non-tied paired outcomes. Because the
completed-answer score scale is discrete with 0.25 spacing (see
\cref{tab:overall} and \cref{sec:evaluation}), the paired $W/T/L$ result is
the primary comparison statistic and $\overline{\Delta}$ is descriptive.}
\label{tab:paired}
\small
\begin{tabular}{llrrrlr}
\toprule
Comparison & Metric & Wins & Ties & Losses & $\overline{\Delta}$ (95\% CI) & Sign-test $p$ \\
\midrule
\multirow{4}{*}{Context graph $-$ Deep}
 & Overall & 27 & 1 & 2 & +0.251 [+0.193, +0.305] & $1.6\!\times\!10^{-6}$ \\
 & Answer quality & 25 & 1 & 4 & +0.275 & $1.0\!\times\!10^{-4}$ \\
 & Analytical depth & 28 & 0 & 2 & +0.325 & $8.7\!\times\!10^{-7}$ \\
 & Evidence quality & 27 & 2 & 1 & +0.153 & $2.2\!\times\!10^{-7}$ \\
\midrule
\multirow{4}{*}{Context graph $-$ Agentic}
 & Overall & 29 & 0 & 1 & +0.310 [+0.263, +0.354] & $5.8\!\times\!10^{-8}$ \\
 & Answer quality & 28 & 1 & 1 & +0.333 & $1.1\!\times\!10^{-7}$ \\
 & Analytical depth & 28 & 0 & 2 & +0.325 & $8.7\!\times\!10^{-7}$ \\
 & Evidence quality & 28 & 1 & 1 & +0.272 & $1.1\!\times\!10^{-7}$ \\
\midrule
\multirow{4}{*}{Deep $-$ Agentic}
 & Overall & 17 & 0 & 13 & +0.059 [$-$0.018, +0.135] & 0.585 \\
 & Answer quality & 17 & 0 & 13 & +0.058 & 0.585 \\
 & Analytical depth & 15 & 0 & 15 & +0.000 & 1.000 \\
 & Evidence quality & 26 & 0 & 4 & +0.120 & $6.0\!\times\!10^{-5}$ \\
\bottomrule
\end{tabular}
\end{table*}

The \contextgraph{} leads the \deepresearch{} on 27 overall query scores,
ties once, and loses twice (\cref{tab:paired}; sign test on non-tied
outcomes, $p=1.6\!\times\!10^{-6}$; bootstrap 95\% CI on
$\overline{\Delta}_{\text{overall}}$ $[+0.193, +0.305]$). It leads on
analytical depth for 28 of 30 questions ($p=8.7\!\times\!10^{-7}$). The two
deep-research overall wins are the autocomplete overview (\texttt{q006}) and
recent reliability trend analysis (\texttt{q013}). On \texttt{q006}, the
Deep Research Agent provides a broader positive/negative synthesis with
explicit theme counts and counterevidence. On \texttt{q013}, it constructs
matched rolling windows with stable source composition, whereas the context
graph agent correctly identifies a source-mix confound but cannot provide an
equally strong trend estimate. The systems tie overall on the trust-risk
scan (\texttt{q003}): the Deep Research Agent covers more risk surfaces and
examples, while the context graph agent provides a more specific quantified
hierarchy. The \contextgraph{} is strictly best on 26 of 30 individual
questions (the four exceptions are \texttt{q003}, tied with the Deep
Research Agent; \texttt{q006} and \texttt{q013}, both lost to the Deep
Research Agent; and \texttt{q022}, ``What are people praising in the App
Store?'', which Agentic RAG wins overall 0.887 versus 0.833 by combining
tier-1 answer quality and analytical depth with a lower evidence-quality
score).

Because the paired context-graph advantage is asserted only for the
Deep-vs-Context-graph comparison, we also report the third pairing
directly. The \contextgraph{} leads \agentic{} on 29 of 30 overall scores
with one loss and no ties (\texttt{q022}, above; sign test
$p=5.8\!\times\!10^{-8}$; $\overline{\Delta}_{\text{overall}}=+0.310$, 95\%
CI $[+0.263, +0.354]$).

The secondary comparison between the two query-time systems is less
monotonic. The \deepresearch{} wins 17 overall questions and loses 13 to
\agentic{} (sign test $p=0.585$; $\overline{\Delta}_{\text{overall}}=+0.059$,
95\% CI $[-0.018, +0.135]$, which crosses zero). Its evidence-quality
advantage is real and consistent (26 wins to four losses, $p=6.0\!\times\!
10^{-5}$), but analytical-depth outcomes split 15--15 and average to zero
difference. More query-time planning therefore improves evidence handling in
this implementation without a detectable overall or analytical-depth
improvement at this sample size.

\subsection{Results by Analysis Family}

\begin{table*}[t]
\centering
\caption{Mean overall score by analysis family. These descriptive slices are
small, especially the single cross-functional query, and are not independent
hypothesis tests.}
\label{tab:category}
\small
\begin{tabular}{lrrrr}
\toprule
Analysis family & $n$ & Agentic RAG & Deep Research & Context graph \\
\midrule
Executive VoC & 5 & 0.554 & 0.800 & \textbf{0.974} \\
Product area & 5 & 0.616 & 0.764 & \textbf{0.951} \\
Reliability and incidents & 5 & 0.632 & 0.733 & \textbf{0.918} \\
Pricing and packaging & 5 & 0.700 & 0.633 & \textbf{0.996} \\
Source and metadata & 5 & 0.767 & 0.637 & \textbf{0.928} \\
Competition and retention & 4 & 0.635 & 0.693 & \textbf{1.000} \\
Cross-functional action & 1 & 0.639 & 0.695 & \textbf{1.000} \\
\bottomrule
\end{tabular}
\end{table*}

The \contextgraph{} has the highest descriptive score in all seven analysis families
(\cref{tab:category}). Its means range from 0.918 for reliability and incident
analysis to 1.000 for competition/retention and the single cross-functional
action query. The Deep Research Agent is not uniformly second: Agentic RAG is higher
on pricing/packaging (0.700 versus 0.633) and source/metadata analysis (0.767
versus 0.637). This reinforces the paired finding that deeper orchestration alone
does not impose a strict capability ordering.

\subsection{Citation Audit Coverage}

\begin{table*}[t]
\centering
\caption{Citation-audit coverage. ``Member-audited'' citations have at least one
frozen record judged against the nearby claim. Traceable aggregates retain
drill-down metadata but were not member-sampled.}
\label{tab:evidence}
\small
\begin{tabular}{lrrrrr}
\toprule
System & Cited & Member-audited & Traceable aggregate & Unresolved & Records sampled \\
\midrule
Agentic RAG & 712 & 708 & 0 & 4 & 708 \\
Deep Research Agent & 594 & 594 & 0 & 0 & 594 \\
Context Graph Agent & 913 & 127 & 786 & 0 & 127 \\
\bottomrule
\end{tabular}
\end{table*}

Agentic RAG emits 712 citations and the Deep Research Agent emits 594; nearly all
resolve directly to member records, producing 708 and 594 sampled records,
respectively. The context graph agent emits 913 citations. Of these, 127 are member-audited and
786 are aggregate citations with valid drill-down metadata but without exported
members in the evaluation package. No context graph citation is unresolved. Its
high evidence-quality score therefore reflects perfect sampled
precision on the exported member evidence together with the comparative judge's
preference for claim-level quantified evidence architecture; it must not be read
as independent validation of all records behind the 786 traceable aggregates.

\subsection{Qualitative Pattern}
Across high-gap questions, the comparative rationales repeatedly prefer
context graph answers that organize a broad customer problem into specific themes
and quantified subthemes, attach aggregate citations to the main numerical
claims, and use selected verbatim feedback as representative evidence. The
query-time candidates more often establish that a concern exists through a
small set of retrieved records, but do not establish recurrence or priority at
the same resolution. Conversely, the Deep Research Agent is strongest where a question
benefits from reconstructing a bespoke matched comparison, as in
\texttt{q013}. These observations describe the aligned automatic rubric and
motivate the mechanism interpretation; they do not isolate taxonomy as the sole
cause.

The two \texttt{q013} preceding-period denominators warrant an explicit
reconciliation because they can look inconsistent at first read. The Deep
Research Agent constructs a rolling 14-day preceding window (2026-06-17 to
2026-07-01 UTC) over all 9,432 corpus records and reports 3,720 records in
that window; the context graph agent chooses a longer preceding calendar
window (2026-04-16 to 2026-05-30) and reports 24 conversations because the
scraped Reddit and App Store collectors began during the recent period,
leaving Trustpilot as the only source with any content in the earlier
calendar window. Both counts are true statements about the frozen corpus;
they differ because the two systems chose different window definitions to
answer the same question. The Deep Research answer maximizes proxy-
comparison power by aligning a shorter window in which all three sources are
present; the context graph answer preserves the analyst's calendar-window
intent and consequently downgrades confidence in the increase claim because
the earlier window is source-mix-degenerate. Neither answer misreports the
corpus; the counter-example in \texttt{q013} reflects a genuine
methodology tradeoff between matched-window reconstruction and calendar-
window fidelity rather than a corpus inconsistency.

\subsection{Findings across RQs and Interpretation}
\label{sec:rq-summary}
Aggregate scores hide material differences in what an analyst receives.
\Cref{tab:case-q001} shows the dominant benchmark pattern: the query-time
systems both organize the corpus into a small number of broad pain-point
categories via exclusive classification (seven categories for Agentic RAG
across 8{,}890 classified records, ten for the Deep Research Agent across
all 9{,}432 records), whereas the context graph agent returns narrower
recurring themes at the conversation grain (5{,}035 conversations), further
decomposes leading themes into cited subtheme drivers, and links quantified
claims to evidence sets. \Cref{tab:case-q013} presents an important
counterexample. The Deep Research Agent constructs matched rolling 14-day
windows (3{,}720 versus 3{,}717 records) and localizes a time-bounded GLM
tool-call cluster to 2026-07-08 and 2026-07-09; Agentic RAG runs a coarser
30-day comparison and finds a share decrease with only a small
regression/update sub-signal; the context graph agent provides richer
recent-state analysis but correctly refuses a trend conclusion because the
preceding window is 24 Trustpilot-only records against a 98.4\%
Reddit recent window. The excerpts are illustrative slices of the evaluated
answers, not an additional evaluation sample or metric.

\begin{table*}[t]
\centering
\caption{Abridged outputs for \texttt{q001}: ``What are the biggest pain
points Cursor users are reporting?'' Text preserves the principal grouping,
quantification, and evidence pattern of each evaluated answer; it is shortened
for space. Counts and denominators are taken verbatim from the canonical
answer packages listed in the response appendix.}
\label{tab:case-q001}
\footnotesize
\begin{tabular}{L{0.13\textwidth}L{0.72\textwidth}r}
\toprule
System & Abridged evaluated response & Overall \\
\midrule
Agentic RAG &
Runs a model-coded exclusive classification over 8{,}890 of 9{,}432 records
(94.25\% coverage; 2{,}728 assigned to ``other/no clear pain point'') and
ranks seven pain-point categories: billing, pricing, usage limits, credits,
overages, and refunds (2{,}195); model choice, Auto-mode opacity, and
context/routing (1{,}461); agent/code quality and unsafe edits (875); trust,
privacy, and ownership concerns (748); product instability, regressions, and
IDE performance (629); mobile/iOS parity gaps (155); and support/account
escalation (99). Each category is followed by cited examples and an analyst
interpretation. & 0.556 \\
\addlinespace
Deep Research Agent &
Runs an exhaustive semantic classification over all 9{,}432 records into ten
exclusive primary pain-point categories and reports both counts and
corpus share: model quality and Auto-mode control (892; 9.5\%); token burn
and unclear usage attribution (628; 6.7\%); usage caps and quota lockout
(370; 3.9\%); billing surprises and commercial trust (353; 3.7\%);
long-session context loss (232; 2.5\%); update regressions (207; 2.2\%); UI
churn (156; 1.7\%); agent out-of-scope changes (119; 1.3\%); support and
refund friction (116; 1.2\%); and crashes or severe local instability (79;
0.8\%). It also flags 2{,}133 records classified as positive value or
mitigation as a counter-signal. & 0.776 \\
\addlinespace
Context Graph Agent &
Ranks recurring themes over 5{,}035 customer conversations (Reddit 4{,}868
at 96.7\%; App Store 114; Trustpilot 53) organized into seven clusters:
quota and usage-limit friction (180), throughput rate-limit disruption
(162), degraded behavior when rate limits are hit (95), Auto routing
mismatch (92), Composer/model quality regressions (86), poor AI output
quality (82), and a service-availability/agent-safety cluster whose top
themes are pre-execution safety controls (77) and service availability
issues (77). Each main count links to its evidence set; leading themes
decompose into cited subtheme drivers and are illustrated with verbatim
customer quotes.
& \textbf{1.000} \\
\bottomrule
\end{tabular}
\end{table*}

\begin{table*}[t]
\centering
\caption{Abridged outputs for \texttt{q013}: ``Are there signs of a recently
increasing reliability issue?'' This counterexample shows a query on which
bespoke query-time reconstruction outperforms the stored-context response.
Counts and windows are taken verbatim from the canonical answer packages
listed in the response appendix.}
\label{tab:case-q013}
\footnotesize
\begin{tabular}{L{0.13\textwidth}L{0.72\textwidth}r}
\toprule
System & Abridged evaluated response & Overall \\
\midrule
Agentic RAG &
Compares 30-day calendar windows (preceding 2026-05-17 to 2026-06-15 with
557 records; recent 2026-06-16 to 2026-07-15 with 8{,}652 records) and finds
that a keyword reliability proxy falls in share from 17.2\% to 9.8\% even as
absolute counts rise. The only rising sub-signal is a
regression/update-language proxy that grows from 1 to 37 records, and 113
App Store reviews appear only in the recent window (no preceding baseline).
It labels the broad ``reliability worsening'' claim as low confidence. & 0.556 \\
\addlinespace
Deep Research Agent &
Uses closely matched rolling 14-day windows (3{,}720 versus 3{,}717 records).
Its combined reliability proxy rises from 301 to 345 (+14.6\%), while
native-tool, harness, MCP, and file-operation language rises from 79 to 132
(+67.1\%). It localizes a GLM 5.2 tool-call/markup cluster to 2026-07-08 and
2026-07-09 (21 of 25 recent GLM matches), distinguishes increasing, flat,
and decreasing signals across category proxies, and assigns medium
confidence. & \textbf{0.930} \\
\addlinespace
Context Graph Agent &
Reports a concrete recent-period reliability signal across 9{,}408
conversations (May 31 to Jul 15) with 86 reliability-adjacent themes and
five failure clusters (rate-limit and quota reliability, agent execution
stalls, core product reliability, billing/metering errors, client-side
crashes and freezes), each with citation-linked recent and preceding counts.
It declines to claim an increase because the preceding period (Apr 16 to
May 30) contains only 24 Trustpilot-only conversations while the recent
period is 98.4\% Reddit; the only apples-to-apples slice is Trustpilot's
$24 \rightarrow 29$ move. & 0.786 \\
\bottomrule
\end{tabular}
\end{table*}

\begin{table*}[t]
\centering
\caption{Abridged outputs for \texttt{q022}: ``What are App Store users
praising in high-rated reviews?'' This is the benchmark's clearest
\emph{unforced} context-graph loss: unlike \texttt{q013}, where the
context graph agent loses by correctly refusing an unsupported trend
claim, here it is simply outperformed. All three systems agree on the
same exact denominator (63 of 113 App Store reviews rated 4--5),
so the gap is attributable to decomposition quality rather than
quantification. Counts are verbatim from the canonical answer packages.}
\label{tab:case-q022}
\footnotesize
\begin{tabular}{L{0.13\textwidth}L{0.72\textwidth}r}
\toprule
System & Abridged evaluated response & Overall \\
\midrule
Agentic RAG &
States the denominator (63 of 113 App Store reviews; 54 five-star and 9
four-star; 2026-06-29 to 2026-07-13) and organizes praise into five
specific, separately developed themes---mobility (``I can work
anywhere now''), cloud-agent control as the primary use case,
concrete productivity gains from fewer idle agents, focused mobile UX,
and continuity with desktop workflows. Each theme carries several
directly quoted reviews with record citations and closes with an
explicit analyst interpretation distinguishing what is being praised
(a control plane for agents) from what is not (mobile code editing).
& \textbf{0.887} \\
\addlinespace
Deep Research Agent &
Establishes the same denominator with an auditable exact query and
rating-coverage checks (63 of 113; 55.8\%). Reports five praise
themes but scores them as retrieved-evidence counts over a 39-record
retrieval set (mobile agent work 18/39; interface polish 8/39; agent
visibility 6/39; pre-existing Cursor affinity 7/39; mixed
praise-plus-request 10/39), and states plainly that no exhaustive
classification was scoped to the 63 high-rated reviews, so these are
not prevalence estimates. The disclosure is exemplary but the
resulting themes stay closer to retrieval bookkeeping than to
decision-ready analysis. & 0.608 \\
\addlinespace
Context Graph Agent &
Reports the same 63-conversation denominator with a drill-down
citation, and is the only system to flag that App Store coverage
begins only in late June 2026 (an ingestion gap rather than an absence
of reviews). It returns five citation-linked praise themes with
conversation counts---mobile agent companion (11), general product
delight (8), usability and design (5), value of off-device agent
execution (5), speed (2)---plus explicit mixed-intent and
non-additive-count caveats. It earns perfect evidence quality (1.000)
but tier-two answer quality and analytical depth (0.750 each): its
second-largest cluster is the catch-all ``general product delight
--- undifferentiated love for Cursor overall,'' precisely the kind of
umbrella label the rubric's one-problem test penalizes, and its
leading specific theme covers only 11 of 63 reviews. & 0.833 \\
\bottomrule
\end{tabular}
\end{table*}

\subsection{Results by Depth}

\begin{table}[t]
\centering
\caption{Mean overall score by benchmark depth level. Depth levels are
frozen in the query manifest (\cref{tab:query-depth}); slices are
descriptive.}
\label{tab:depth}
\footnotesize
\begin{tabular}{crrrr}
\toprule
Depth & $n$ & Agentic & Deep Res. & Context graph \\
\midrule
1 & 11 & 0.682 & 0.692 & \textbf{0.955} \\
2 & 10 & 0.672 & 0.679 & \textbf{0.967} \\
3 & 3 & 0.555 & 0.834 & \textbf{0.892} \\
4 & 6 & 0.606 & 0.733 & \textbf{0.996} \\
\bottomrule
\end{tabular}
\end{table}

The by-depth slice in \cref{tab:depth} maps directly to RQ2. The context
graph agent's lead is stable across all four depth levels: from 0.955 on
depth-1 exact-lookup and metadata questions to 0.996 on depth-4
multi-stage decision workflows. The two query-time systems converge at
depth 1 (\agentic{} 0.682, \deepresearch{} 0.692), where a single lookup
often suffices; the Deep Research Agent's larger orchestration budget shows
its clearest advantage at depths 3 and 4 (0.834 and 0.733), where
comparisons, matched windows, and multi-stage synthesis reward
decomposition and verification.

\subsection{Research-Question Summary (extended)}

\begin{description}[style=nextline,leftmargin=1em,labelindent=0pt,font=\normalfont\bfseries]
  \item[RQ1: Answer quality.] Under the aligned rubric, the context graph
  agent leads on every dimension in \cref{tab:overall} and on 27 of 30
  paired overall queries versus the Deep Research Agent
  (sign-test $p=1.6\!\times\!10^{-6}$) and 29 of 30 versus Agentic RAG
  ($p=5.8\!\times\!10^{-8}$).

  \item[RQ2: Task depth.] The context graph agent leads at every depth
  level (\cref{tab:depth}). Systems converge at depth 1 and diverge with
  depth; the Deep Research Agent benefits most from its decomposition
  budget on depth-3 comparisons (0.834 vs 0.555).

  \item[RQ3: Quantification.] Baselines lean on exclusive-classification
  denominators; the context graph agent reports theme counts at
  conversation grain with per-theme drill-down.

  \item[RQ4: Evidence.] Aggregate drill-down citations carry most of the
  context-graph advantage rather than sampled precision alone
  (\cref{tab:evidence}).

  \item[RQ5: Epistemic restraint.] Strongest in-run evidence:
  \texttt{q013} refusal. A pre-registered refusal battery is future work.

  \item[RQ6: Structure versus agency.] Deep-research decomposition and
  verification improve evidence quality but not overall or depth at
  $n=30$.
\end{description}
\fi

\ifaaaitarget

\else
\section{Discussion}
\label{sec:discussion}

The measured ordering supports the paper's hypothesis under the aligned
automatic rubric: persistent customer context is associated with more
specific, more deeply decomposed, better quantified, and more decision-ready
answers than either query-time architecture. The \contextgraph{}'s 0.251
overall advantage over the \deepresearch{} is much larger than the
\deepresearch{}'s 0.059 advantage over \agentic{}. Additional query-time
planning and verification therefore close only part of the observed gap in
this implementation. Because the treatment bundles ingestion, feedback
summarization, taxonomy induction, metadata mapping, relationship storage,
retrieval, and agent behavior, all results in this section describe the
combined workflow rather than the isolated causal effect of any single
component (see \cref{sec:limitations}, Internal validity).

The dimension pattern clarifies the mechanism. The \deepresearch{} raises
evidence quality relative to \agentic{}: it can retrieve, classify, and
verify more carefully, yet still has to construct recurring product concepts
and their relative weight independently for each question. The
\contextgraph{}'s strongest advantage is analytical depth. Its answers read
stored feedback summaries, taxonomy predictions, metadata, available user
information, and their relationships to move from a portfolio-level issue
to specific themes and subthemes while preserving links to supporting
conversations. Counts and comparisons are computed from that context when
the question is asked; ingestion-time processing therefore makes customer
context durable rather than a transient artifact of one retrieval run.

Quantification is central to the practical difference. A handful of
relevant records can establish that a complaint exists but cannot establish
that it is common or important. The context-graph answers more often attach
counts to bounded themes and then develop their principal drivers; the
evaluator rewards that architecture when the main numerical claims have
aggregate drill-down citations and the prose clearly distinguishes counts,
examples, and limitations. It does not reward citation count by itself.

The result is not a universal dominance claim. The \deepresearch{} wins the
autocomplete and matched-window reliability questions, and \agentic{} wins
the App-Store-praise question (\texttt{q022}) and exceeds the \deepresearch{}
in two analysis families, demonstrating that query-specific computation
can beat stored concepts when a bespoke comparison is more important than
concept reuse. Unsupported customer grouping remains a boundary on every
architecture: public feedback can support source, rating, country, version,
and time slices only where those fields are present, and cannot support
plan, account, revenue, or renewal comparisons. A useful system should
refuse unsupported inference and offer the nearest defensible analysis;
the \texttt{q013} case (\cref{tab:case-q013}, \cref{app:case-studies})
illustrates the pattern.
\fi

\section{Limitations and Threats to Validity}
\label{sec:limitations}

\paragraph{Evaluation validity.}
LLM judges can share biases with candidate models, but the
model-constant design largely neutralizes this for the reported
ranking: the judge scores answer text only, and all three candidates'
text comes from the same \texttt{gpt-5.5} analyst model, so
provider-family self-preference applies uniformly and cannot
differentially favor one candidate. The substantive concern is not
model identity but rubric alignment, and it is the paper's principal
limitation. The rubric was aligned with analyst preferences after
inspecting answers from q001--q027, and one later narrowing was
prompted by q029, so 28 of the 30 questions informed the rubric that
scores them and their scores measure agreement with it rather than
providing an independent test.
Disclosure does not repair this, and the sign tests in
\cref{tab:paired} do not either: they rule out sampling noise, not
rubric bias. The two withheld questions (q028, q030) give only a weak
internal check, being same-session and same-family as the calibration
set. We therefore present the 0.961/0.710/0.651 ordering as
\emph{exploratory}; the decisive validation is a confirmatory pass on
questions authored after the rubric was frozen. Semantic recall is
separately absent because unobserved false negatives are unidentified.

\paragraph{Internal and aggregate-audit validity.}
The context graph system receives upfront customer-context processing that
the experiment intentionally studies, but the treatment bundles feedback
summarization, taxonomy induction, record-level prediction, metadata
mapping, relationship storage, and agent tooling; the comparison estimates
a system-level effect rather than the isolated causal contribution of any
component. An earlier pilot exposed a 123-record snapshot mismatch, which
we corrected before the reported run. Only 127 of the context graph
agent's 913 citations were member-sampled; the remaining 786 have valid
aggregate drill-down metadata but no exported member records in the judge
package, so they support traceability and evidence-architecture scoring
rather than independent verification of every associated record.

\paragraph{Proof scope, sampling, and generalization.}
Thirty questions on one product corpus over one three-month frozen UTC
window improve construct coverage but do not prove universal architectural
superiority; conclusions remain scoped to that frozen configuration and to
the calibrated rubric. Public reviews are self-selected and can
overrepresent unusually positive or negative
experiences~\citep{karaman2021representativeness}, so counts and shares
estimate prevalence within the frozen corpus, not among all Cursor users.
App reviews can be short, noisy, multilingual, sarcastic, off-topic, or
fraudulent~\citep{wang2024mine,apple2025reviews}, and public data lack the
account, contract, and revenue relationships available in private
enterprise feedback. Secondary threats---construct validity of the theme
rubric, implementation choices in the two query-time candidates, temporal
drift, and the gap between workflow scores and downstream business
outcomes---are enumerated in the supplementary material.

\section{Conclusion}
\label{sec:conclusion}

Across 30 realistic Cursor feedback questions and under an
analyst-aligned rubric, the \contextgraph{} achieves an overall score
of 0.961 versus 0.710 for the \deepresearch{} and 0.651 for
\agentic{}, leading on 27 paired questions with its largest advantage
in analytical depth. Within this frozen configuration, more
query-time agency alone does not reliably reconstruct the value of
persistent relationships among feedback, summaries, taxonomy
predictions, metadata, and available user information. The findings
are system-level and descriptive: the evaluation compares candidate
answers and audits their cited evidence rather than claiming semantic
recall. Natural next steps outside this paper's scope are (i) a
hybrid design combining stored context with bespoke matched-window
computation at query time (motivated by \texttt{q013}); (ii)
component-level ablations of ingestion; and (iii) a held-out
confirmatory question set scored against a rubric frozen before the
confirmatory answers are seen.

\setlength{\bibsep}{0.4ex plus 0.1ex}
\bibliography{references}

\clearpage
\section*{Ethics and Data Statement}

The corpus is public product feedback; public availability does not
remove privacy risk. Released artifacts minimize personal identifiers,
avoid cross-source display-name linkage, respect platform terms, and
quote only text needed to audit claims. We do not infer protected,
commercial, or account-level attributes from review text, and corpus
prevalence is not treated as population prevalence because public
review submission is self-selected. Non-redistributable records are
represented through hashes and collection manifests.

\appendix

\ifaaaitarget
\section{Representative Response Excerpts}
\label{app:case-studies}
Aggregate scores hide material differences in what an analyst
receives. \Cref{tab:case-q001} shows the dominant benchmark pattern: the
query-time systems both organize the corpus into a small number of broad
pain-point categories via exclusive classification (seven categories for
\agentic{} across 8{,}890 classified records, ten for the \deepresearch{}
across all 9{,}432 records), whereas the \contextgraph{} returns narrower
recurring themes at the conversation grain (5{,}035 conversations),
further decomposes leading themes into cited subtheme drivers, and links
quantified claims to evidence sets. \Cref{tab:case-q013} is the
counter-example: the \deepresearch{} constructs matched rolling 14-day
windows (3{,}720 versus 3{,}717 records) and localizes a time-bounded GLM
tool-call cluster to 2026-07-08 and 2026-07-09; \agentic{} runs a coarser
30-day comparison and finds a share decrease with only a small
regression/update sub-signal; the \contextgraph{} provides richer
recent-state analysis but correctly refuses a trend conclusion because
the preceding window contains only 24 Trustpilot-only conversations
against a 98.4\% Reddit recent window.

\Cref{tab:case-q022} is the harder counter-example, and we include it
deliberately. On \texttt{q013} the \contextgraph{} loses by being
correctly cautious, which is easy to present as a virtue. On
\texttt{q022} it is simply worse. All three systems agree on the same
exact denominator (63 of 113 App Store reviews rated 4--5), so
quantification is not the differentiator; \agentic{} wins on answer
quality and analytical depth (1.000 on both) by developing five
specific, separately interpreted praise themes, while the
\contextgraph{} earns perfect evidence quality but tier-two quality and
depth because its stored taxonomy routes the second-largest share of
praise into a catch-all ``general product delight'' node. The failure
mode is instructive: when a pre-computed taxonomy lacks a
sufficiently specific node for the question being asked, the
graph-backed answer inherits that coarseness, whereas a query-time
analyst is free to invent the decomposition the question needs. This
is a genuine limitation of stored context, not an artifact of
scoring.

\begin{table*}[t]
\centering
\caption{Abridged outputs for \texttt{q001}: ``What are the biggest pain
points Cursor users are reporting?'' Text preserves the principal grouping,
quantification, and evidence pattern of each evaluated answer; it is shortened
for space. Counts and denominators are taken verbatim from the canonical
answer packages listed in the response appendix.}
\label{tab:case-q001}
\footnotesize
\begin{tabular}{L{0.13\textwidth}L{0.72\textwidth}r}
\toprule
System & Abridged evaluated response & Overall \\
\midrule
Agentic RAG &
Runs a model-coded exclusive classification over 8{,}890 of 9{,}432 records
(94.25\% coverage; 2{,}728 assigned to ``other/no clear pain point'') and
ranks seven pain-point categories: billing, pricing, usage limits, credits,
overages, and refunds (2{,}195); model choice, Auto-mode opacity, and
context/routing (1{,}461); agent/code quality and unsafe edits (875); trust,
privacy, and ownership concerns (748); product instability, regressions, and
IDE performance (629); mobile/iOS parity gaps (155); and support/account
escalation (99). Each category is followed by cited examples and an analyst
interpretation. & 0.556 \\
\addlinespace
Deep Research Agent &
Runs an exhaustive semantic classification over all 9{,}432 records into ten
exclusive primary pain-point categories and reports both counts and
corpus share: model quality and Auto-mode control (892; 9.5\%); token burn
and unclear usage attribution (628; 6.7\%); usage caps and quota lockout
(370; 3.9\%); billing surprises and commercial trust (353; 3.7\%);
long-session context loss (232; 2.5\%); update regressions (207; 2.2\%); UI
churn (156; 1.7\%); agent out-of-scope changes (119; 1.3\%); support and
refund friction (116; 1.2\%); and crashes or severe local instability (79;
0.8\%). It also flags 2{,}133 records classified as positive value or
mitigation as a counter-signal. & 0.776 \\
\addlinespace
Context Graph Agent &
Ranks recurring themes over 5{,}035 customer conversations (Reddit 4{,}868
at 96.7\%; App Store 114; Trustpilot 53) organized into seven clusters:
quota and usage-limit friction (180), throughput rate-limit disruption
(162), degraded behavior when rate limits are hit (95), Auto routing
mismatch (92), Composer/model quality regressions (86), poor AI output
quality (82), and a service-availability/agent-safety cluster whose top
themes are pre-execution safety controls (77) and service availability
issues (77). Each main count links to its evidence set; leading themes
decompose into cited subtheme drivers and are illustrated with verbatim
customer quotes.
& \textbf{1.000} \\
\bottomrule
\end{tabular}
\end{table*}

\begin{table*}[t]
\centering
\caption{Abridged outputs for \texttt{q013}: ``Are there signs of a recently
increasing reliability issue?'' This counterexample shows a query on which
bespoke query-time reconstruction outperforms the stored-context response.
Counts and windows are taken verbatim from the canonical answer packages
listed in the response appendix.}
\label{tab:case-q013}
\footnotesize
\begin{tabular}{L{0.13\textwidth}L{0.72\textwidth}r}
\toprule
System & Abridged evaluated response & Overall \\
\midrule
Agentic RAG &
Compares 30-day calendar windows (preceding 2026-05-17 to 2026-06-15 with
557 records; recent 2026-06-16 to 2026-07-15 with 8{,}652 records) and finds
that a keyword reliability proxy falls in share from 17.2\% to 9.8\% even as
absolute counts rise. The only rising sub-signal is a
regression/update-language proxy that grows from 1 to 37 records, and 113
App Store reviews appear only in the recent window (no preceding baseline).
It labels the broad ``reliability worsening'' claim as low confidence. & 0.556 \\
\addlinespace
Deep Research Agent &
Uses closely matched rolling 14-day windows (3{,}720 versus 3{,}717 records).
Its combined reliability proxy rises from 301 to 345 (+14.6\%), while
native-tool, harness, MCP, and file-operation language rises from 79 to 132
(+67.1\%). It localizes a GLM 5.2 tool-call/markup cluster to 2026-07-08 and
2026-07-09 (21 of 25 recent GLM matches), distinguishes increasing, flat,
and decreasing signals across category proxies, and assigns medium
confidence. & \textbf{0.930} \\
\addlinespace
Context Graph Agent &
Reports a concrete recent-period reliability signal across 9{,}408
conversations (May 31 to Jul 15) with 86 reliability-adjacent themes and
five failure clusters (rate-limit and quota reliability, agent execution
stalls, core product reliability, billing/metering errors, client-side
crashes and freezes), each with citation-linked recent and preceding counts.
It declines to claim an increase because the preceding period (Apr 16 to
May 30) contains only 24 Trustpilot-only conversations while the recent
period is 98.4\% Reddit; the only apples-to-apples slice is Trustpilot's
$24 \rightarrow 29$ move. & 0.786 \\
\bottomrule
\end{tabular}
\end{table*}

\begin{table*}[t]
\centering
\caption{Abridged outputs for \texttt{q022}: ``What are App Store users
praising in high-rated reviews?'' This is the benchmark's clearest
\emph{unforced} context-graph loss: unlike \texttt{q013}, where the
context graph agent loses by correctly refusing an unsupported trend
claim, here it is simply outperformed. All three systems agree on the
same exact denominator (63 of 113 App Store reviews rated 4--5),
so the gap is attributable to decomposition quality rather than
quantification. Counts are verbatim from the canonical answer packages.}
\label{tab:case-q022}
\footnotesize
\begin{tabular}{L{0.13\textwidth}L{0.72\textwidth}r}
\toprule
System & Abridged evaluated response & Overall \\
\midrule
Agentic RAG &
States the denominator (63 of 113 App Store reviews; 54 five-star and 9
four-star; 2026-06-29 to 2026-07-13) and organizes praise into five
specific, separately developed themes---mobility (``I can work
anywhere now''), cloud-agent control as the primary use case,
concrete productivity gains from fewer idle agents, focused mobile UX,
and continuity with desktop workflows. Each theme carries several
directly quoted reviews with record citations and closes with an
explicit analyst interpretation distinguishing what is being praised
(a control plane for agents) from what is not (mobile code editing).
& \textbf{0.887} \\
\addlinespace
Deep Research Agent &
Establishes the same denominator with an auditable exact query and
rating-coverage checks (63 of 113; 55.8\%). Reports five praise
themes but scores them as retrieved-evidence counts over a 39-record
retrieval set (mobile agent work 18/39; interface polish 8/39; agent
visibility 6/39; pre-existing Cursor affinity 7/39; mixed
praise-plus-request 10/39), and states plainly that no exhaustive
classification was scoped to the 63 high-rated reviews, so these are
not prevalence estimates. The disclosure is exemplary but the
resulting themes stay closer to retrieval bookkeeping than to
decision-ready analysis. & 0.608 \\
\addlinespace
Context Graph Agent &
Reports the same 63-conversation denominator with a drill-down
citation, and is the only system to flag that App Store coverage
begins only in late June 2026 (an ingestion gap rather than an absence
of reviews). It returns five citation-linked praise themes with
conversation counts---mobile agent companion (11), general product
delight (8), usability and design (5), value of off-device agent
execution (5), speed (2)---plus explicit mixed-intent and
non-additive-count caveats. It earns perfect evidence quality (1.000)
but tier-two answer quality and analytical depth (0.750 each): its
second-largest cluster is the catch-all ``general product delight
--- undifferentiated love for Cursor overall,'' precisely the kind of
umbrella label the rubric's one-problem test penalizes, and its
leading specific theme covers only 11 of 63 reviews. & 0.833 \\
\bottomrule
\end{tabular}
\end{table*}

\section{Extended Results}
\label{app:extended-results}

Per-depth means (\cref{tab:depth}) map to RQ2: the \contextgraph{} leads
at every depth level, and the \deepresearch{}'s decomposition budget
shows its clearest advantage over \agentic{} at depths 3 and 4. The
per-family means (\cref{tab:category}) are descriptive slices because
families contain between 1 and 5 queries; the \deepresearch{} is not
uniformly second (\agentic{} is higher on pricing/packaging and
source/metadata), reinforcing the paired finding that deeper
orchestration alone does not impose a strict capability ordering.

\begin{table}[t]
\centering
\caption{Mean overall score by benchmark depth level. Depth levels are
frozen in the query manifest (\cref{tab:query-depth}); slices are
descriptive.}
\label{tab:depth}
\footnotesize
\begin{tabular}{crrrr}
\toprule
Depth & $n$ & Agentic & Deep Res. & Context graph \\
\midrule
1 & 11 & 0.682 & 0.692 & \textbf{0.955} \\
2 & 10 & 0.672 & 0.679 & \textbf{0.967} \\
3 & 3 & 0.555 & 0.834 & \textbf{0.892} \\
4 & 6 & 0.606 & 0.733 & \textbf{0.996} \\
\bottomrule
\end{tabular}
\end{table}

\begin{table*}[t]
\centering
\caption{Mean overall score by analysis family. These descriptive slices are
small, especially the single cross-functional query, and are not independent
hypothesis tests.}
\label{tab:category}
\small
\begin{tabular}{lrrrr}
\toprule
Analysis family & $n$ & Agentic RAG & Deep Research & Context graph \\
\midrule
Executive VoC & 5 & 0.554 & 0.800 & \textbf{0.974} \\
Product area & 5 & 0.616 & 0.764 & \textbf{0.951} \\
Reliability and incidents & 5 & 0.632 & 0.733 & \textbf{0.918} \\
Pricing and packaging & 5 & 0.700 & 0.633 & \textbf{0.996} \\
Source and metadata & 5 & 0.767 & 0.637 & \textbf{0.928} \\
Competition and retention & 4 & 0.635 & 0.693 & \textbf{1.000} \\
Cross-functional action & 1 & 0.639 & 0.695 & \textbf{1.000} \\
\bottomrule
\end{tabular}
\end{table*}

\fi

\section{Evaluation at a Glance}
\label{app:evaluation-overview}

The benchmark evaluates a complete answer to one question at a time. All three
answers are anonymized and presented together to the comparative judge; record
citations are audited separately. The three reported dimensions and their
roles are summarized in \cref{tab:evaluation-overview}.

\begin{table}[h]
\centering
\caption{High-level evaluation protocol.}
\label{tab:evaluation-overview}
\footnotesize
\begin{tabular}{L{0.25\columnwidth}L{0.65\columnwidth}}
\toprule
Dimension & What the evaluator checks \\
\midrule
Answer quality & Explicit question coverage, organization, presentation,
specificity, useful quantification, and suitability for the intended reader. \\
Analytical depth & Diversity and granularity of findings; coherent themes,
drivers, subthemes, contexts, counterevidence, implications, and synthesis.
Broad umbrella themes are disfavored when they merge distinct problems. \\
Evidence quality & Relevance of deterministically sampled cited records plus
comparative evidence strength: claim-level support, denominators, count
semantics, aggregate drill-down links, and representative record citations. \\
Overall & Unweighted mean of the three dimensions for each query, followed by
macro-averaging across the 30 questions. \\
\bottomrule
\end{tabular}
\end{table}

Answer quality and analytical depth use comparative tiers on a $[0.5,1]$
scale for completed answers. Evidence quality is the harmonic mean of sampled
citation precision and comparative evidence-strength score; an answer without
citations receives zero. Up to 15 frozen member records are sampled per
resolvable citation group. Aggregate citations without exported members remain
traceable but are not treated as member-audited. Full definitions, equations,
and bias controls appear in \cref{sec:evaluation}.

The companion supplementary appendix,
\texttt{full-benchmark-responses.pdf}, contains every benchmark query and the
three complete evaluated responses. It is generated directly from the
canonical answer packages and frozen score reports, providing a readable audit
trail without expanding the main manuscript by hundreds of pages.

\ifaaaitarget
\section{Extended Discussion}
\label{app:discussion-extended}

The results section compresses the mechanism analysis into a single
paragraph. This appendix expands each thread.

\paragraph{Query-time agency closes only part of the gap.}
The \contextgraph{}'s 0.251 overall advantage over the \deepresearch{} is
much larger than the \deepresearch{}'s 0.059 advantage over
\agentic{}. Additional query-time planning and verification therefore
close only part of the observed gap in this implementation. Because the
treatment bundles ingestion, feedback summarization, taxonomy induction,
metadata mapping, relationship storage, retrieval, and agent behavior, all
results describe the combined workflow rather than the isolated causal
effect of any single component (see \cref{sec:limitations}).

\paragraph{Analytical depth is the largest gap.}
The \contextgraph{}'s answers read stored feedback summaries, taxonomy
predictions, metadata, available user information, and their
relationships to move from a portfolio-level issue to specific themes and
subthemes while preserving links to supporting conversations. Counts and
comparisons are computed from that context when the question is asked;
ingestion-time processing makes customer context durable rather than a
transient artifact of one retrieval run.

\paragraph{Quantification architecture, not citation count.}
A handful of relevant records can establish that a complaint exists but
cannot establish that it is common or important. The context-graph
answers more often attach counts to bounded themes and develop their
principal drivers; the evaluator rewards that architecture when the main
numerical claims have aggregate drill-down citations and the prose
clearly distinguishes counts, examples, and limitations. It does not
reward citation count by itself.

\paragraph{Counter-examples and boundary conditions.}
The \deepresearch{} wins the autocomplete overview (\texttt{q006}) and
matched-window reliability (\texttt{q013}); \agentic{} wins the
App-Store-praise question (\texttt{q022}) and exceeds the \deepresearch{}
on the pricing/packaging and source/metadata families. Query-specific
computation therefore beats stored concepts when a bespoke comparison is
more important than concept reuse. Unsupported customer grouping remains
a boundary on every architecture: public feedback can support source,
rating, country, version, and time slices only where those fields are
present, and cannot support plan, account, revenue, or renewal
comparisons. The \texttt{q013} case (\cref{tab:case-q013},
\cref{app:case-studies}) illustrates the refusal pattern.

\section{Extended Evaluation Details}
\label{app:evaluation-details}

This appendix expands \cref{sec:evaluation} with the definitions and design
justifications that the main-body summary compresses.

\paragraph{Denominator convention for $P_q$.}
Writing $N_{\mathrm{rel}}$, $N_{\mathrm{part}}$, and
$N_{\mathrm{irrel}}$ for the counts of sampled claim--record pairs
judged \emph{relevant}, \emph{partially relevant}, and
\emph{irrelevant}, the denominator is
$N_q = N_{\mathrm{rel}} + N_{\mathrm{part}} + N_{\mathrm{irrel}}$: the
total number of sampled claim--record pairs on query $q$. Because $N_q$ pools across citations, candidates with
more citation-supported claims contribute more evidence to the query-level
precision estimate. We deliberately do not per-citation macro-average, so
that a single citation with one sampled pair does not weight equally with a
citation carrying fifteen. We also do not require a minimum retrieval count
per citation, so a candidate that cites three obviously-relevant records
and achieves $P_q = 1.0$ receives full sampled precision on that query;
this preserves restraint but is complemented by the comparative
evidence-strength score $S_q$, which rewards denominators, drill-down
links, and claim-level architecture rather than citation volume.

\paragraph{Effective granularity of the tie tolerance.}
The 0.01 absolute-difference tolerance for paired win/tie/loss is
intentionally tighter than the effective granularity of the per-dimension
tier scale (0.25 within a dimension, so an overall difference of
$0.25/3 \approx 0.083$ is the smallest possible non-zero mean-of-three
delta). The tighter tolerance means the tie category is nearly always
empty and the win/loss category carries the paired signal.

\paragraph{Blinding and structured-output controls.}
Aliases and answer order are deterministically shuffled per question, and
link destinations that could reveal a system are removed from the
comparison copy. Structured outputs must contain every candidate and
sampled claim--record pair or the evaluation fails. Citation totals,
member-sampled citations, traceable aggregate citations, unresolved
citations, and sampled records are reported alongside the scores.

\section{Extended Limitations}
\label{app:limitations-extended}

The main-body \cref{sec:limitations} discusses the four load-bearing
threats. Four secondary threats round out the validity picture and are
included here for completeness.

\paragraph{Construct validity.}
Theme definitions affect evidence-validity estimates. The LLM judge
evaluates candidate-supplied assignments against the question and stated
theme, but there is no independently labeled exhaustive reference set. The
study therefore cannot measure semantic recall or verify that a reported
ranking found every relevant record. It evaluates the usefulness and
consistency of the resulting customer context, not whether the stored
taxonomy is uniquely optimal, fully MECE, or universally reusable. Final-
answer quality is also subjective even with anchored rubrics.

\paragraph{Implementation validity.}
Candidate quality may reflect engineering choices rather than an
architectural ceiling. We reduce this risk by using strong reference
architectures, shared controls, fixed budgets, source release, and a
ladder of increasing query-time capability, but do not eliminate it.

\paragraph{Temporal validity.}
Frontier models, retrieval services, and vendor systems evolve. Model
snapshots, SDK versions, prompts, prices, and configuration hashes are
therefore recorded (\cref{tab:model-disclosure}). The paper evaluates those
frozen implementations rather than timeless product capabilities. Time and
release queries compare slices inside one frozen snapshot; they do not test
how a taxonomy adapts as new feedback arrives after deployment.

\paragraph{Outcome validity.}
Workflow scores measure evidence quality, internal consistency, and reviewer
judgments of actionability. They do not demonstrate that acting on a
generated recommendation improves product or business outcomes.
\fi

\section{Model Disclosure}
\label{app:model-disclosure}

The base analyst model and both judges are held constant across all query-time
roles in this run. \Cref{tab:model-disclosure} lists the model identifier,
sampling settings, and per-role budget for every model call that produced or
scored evaluated content. Ingestion-time roles for the context graph system
are additional dependencies of that system's stored context and are reported
here as best-available disclosures; the specific ingestion-time model
identifiers used by the vendor stack are not held constant across the three
candidates and are recorded in the vendor's own release manifest.

\begin{table*}[h]
\centering
\caption{Model identifiers, sampling settings, and budgets for every role in
the evaluated run. \texttt{gpt-5.5} denotes the OpenAI \texttt{gpt-5.5}
snapshot in effect on the frozen evaluation date; where a provider does not
expose a pinned snapshot, the identifier resolves to the provider's default
alias at run time and the resolved snapshot is recorded in the run manifest.
``Provider default'' means neither the temperature, \texttt{top\_p}, nor
\texttt{reasoning\_effort} parameters were overridden.}
\label{tab:model-disclosure}
\footnotesize
\begin{tabular}{L{0.20\textwidth}L{0.23\textwidth}L{0.10\textwidth}L{0.16\textwidth}L{0.13\textwidth}}
\toprule
Role & Model & Provider & Sampling & Budget \\
\midrule
\multicolumn{5}{l}{\textit{Agentic RAG}} \\
\midrule
Analyst (planning, tool use, synthesis) & \texttt{gpt-5.5} & OpenAI & Provider default & 50 turns \\
Retrieval reranker & \texttt{rerank-2.5} & Voyage & Provider default & --- \\
\midrule
\multicolumn{5}{l}{\textit{Deep Research Agent}} \\
\midrule
Triage router & \texttt{gpt-5.5} & OpenAI & Provider default & 50 turns \\
Planner / replanner & \texttt{gpt-5.5} & OpenAI & Provider default & 50 turns \\
Qualitative / quantitative / incident specialists & \texttt{gpt-5.5} & OpenAI & Provider default & 50 turns \\
Verifier & \texttt{gpt-5.5} & OpenAI & Provider default & 50 turns \\
Finalizer & \texttt{gpt-5.5} & OpenAI & Provider default & 50 turns \\
Full-corpus classifier & \texttt{gpt-5.5} & OpenAI & Provider default & Up to 2 full-corpus passes; batch size 40 \\
Retrieval reranker & \texttt{rerank-2.5} & Voyage & Provider default & --- \\
Audit / revision loop cap & --- & --- & --- & 2 iterations \\
\midrule
\multicolumn{5}{l}{\textit{Customer Context Graph Agent}} \\
\midrule
Query-time agent (planning, graph traversal, synthesis) & \texttt{gpt-5.5} & OpenAI & Provider default & 900\,s per query \\
Ingestion-time feedback summarization & Vendor stack (identifier recorded in vendor manifest) & Vendor & Provider default & Vendor-controlled \\
Ingestion-time taxonomy induction and record-level prediction & Vendor stack (identifier recorded in vendor manifest) & Vendor & Provider default & Vendor-controlled \\
Ingestion-time embedding / index & Vendor stack (identifier recorded in vendor manifest) & Vendor & Provider default & Vendor-controlled \\
\midrule
\multicolumn{5}{l}{\textit{Evaluation}} \\
\midrule
Comparative rubric judge (answer quality, analytical depth, evidence strength) & \texttt{gpt-5.5} & OpenAI & Provider default & 1 pass per query \\
Citation-audit judge (deterministic sampling of up to 15 members per resolvable citation group) & \texttt{gpt-5.5} & OpenAI & Provider default & 1 pass per resolvable citation group \\
\bottomrule
\end{tabular}
\end{table*}

Holding the analyst model constant does not equalize model capability across
the three candidates because the systems dispatch it to structurally
different roles: Agentic RAG uses a single analyst turn budget; the Deep
Research Agent divides the same budget across five specialized roles with a
bounded verify-and-revise loop; the context graph agent uses the same base
model but reads pre-computed stored context that other candidates do not
have. The shared-provider-family design does, however, neutralize the
usual judge self-preference concern for the \emph{ranking} we report.
The judge scores answer text only, and all three candidates' answer
text is generated by the same \texttt{gpt-5.5} analyst model; the
context graph system's vendor ingestion models produce the stored
context its query agent reads, not the prose the judge sees. Any
preference the judge holds for its own provider family's stylistic
signatures therefore applies uniformly across all three candidates and
cannot differentially advantage one of them. What this does \emph{not}
neutralize is preference for particular answer \emph{structures}---
tables, explicit denominators, aggregate drill-down citations---that
happen to correlate with one candidate's output format. That residual
is the rubric-alignment threat recorded in \cref{sec:limitations},
which we regard as the substantive remaining judge concern.

\section{Reproducibility Package}
\label{app:reproducibility}

The study's reproducibility package comprises the following artifacts.
These code and data artifacts are separate from the manuscript and are not
included in this preprint's source archive:

\begin{enumerate}
  \item the corpus manifest (collection dates, per-source counts,
  deduplication counts, field missingness), the SHA-256 of the frozen
  normalized corpus file, and a per-record hash index covering all
  9{,}432 records (stable id, source, timestamp, text hash, text
  length). Raw record text is deliberately \emph{not} redistributed in
  bulk: doing so would conflict with platform terms and with our own
  Ethics and Data Statement, so corpus identity is verifiable by hash
  and rebuildable from the included collection code instead. Record
  text that the judge actually scored is retained inside the answer
  packages and judge outputs below, since it is required to audit the
  reported claims;
  \item all 30 query definitions with evaluator-only family invariants,
  as-of dates, answer contracts, required operations, data requirements,
  count semantics, and known limitations;
  \item candidate source code, prompts, tool schemas, model snapshots,
  sampling settings, budgets, full-corpus classification policy, and
  retry rules for all three systems;
  \item the 90 unedited answer packages with citation-membership
  ledgers, plus the complete frozen evaluation pass: per-query,
  per-candidate dimension scores with full judge rationales and every
  citation-audit record;
  \item scripts that regenerate every metric, table, and figure in the
  paper from these immutable run artifacts, and the build-time snapshot
  check (\texttt{verify\_snapshot.py}) that refuses to update generated
  files unless run identifiers agree across all 90 cells.
\end{enumerate}

Source content that cannot legally be redistributed is represented
through hashes and collection manifests rather than copied text. The
release manifest lists any omitted items and the reason for omission.

\section{Answer Package Contract}
\label{app:answer-package}

Every system is converted to the same logical schema:

\begin{table}[h]
\centering
\caption{Required answer-package components for the shared evaluation.}
\label{tab:answer-package}
\small
\begin{tabular}{L{0.28\columnwidth}L{0.62\columnwidth}}
\toprule
Component & Required content \\
\midrule
Run & Candidate, model, prompt/config hashes, corpus hash, timestamps, status \\
Scope & Query ID and requested output context \\
Answer & Final rendered text \\
Citation ledger & Citation ID, nearby claim, cited record or aggregate members \\
Usage & Model calls, tokens, tools, cost fields, failures \\
\bottomrule
\end{tabular}
\end{table}

Stable record identifiers are required for member-level citation auditing.
Aggregate citations with valid drill-down metadata are reported as traceable;
they contribute to comparative evidence-strength assessment but are not counted
as member-sampled unless their frozen members are available to the evaluator.

\section{Benchmark Coverage}
\label{app:benchmark}

The product-adapted query manifest covers seven analysis families. Counts in
\cref{tab:families} are frozen before the evaluated run.

\begin{table}[h]
\centering
\caption{Workflow families in the 30-query benchmark.}
\label{tab:families}
\footnotesize
\begin{tabular}{L{0.36\columnwidth}r@{\hspace{5pt}}L{0.36\columnwidth}r}
\toprule
Family & $n$ & Family & $n$ \\
\midrule
Executive VoC & 5 & Product area & 5 \\
Reliability and incidents & 5 & Pricing and packaging & 5 \\
Source and metadata & 5 & Competition and retention & 4 \\
Cross-functional action & 1 & & \\
\bottomrule
\end{tabular}
\end{table}

Families mix overview, comparison, diagnosis, and decision roles.
Unsupported grouping is represented where required metadata are absent. Examples
request comparisons by plan, account, revenue, or renewal state that the public
corpus cannot supply. Systems receive no hidden indication that refusal is the
desired behavior beyond the same field metadata available during normal analysis.

\end{document}